%% file: main.tex
\documentclass[10pt,twocolumn,letterpaper]{article}
\usepackage{times}
\usepackage{epsfig}
\usepackage{graphicx}
\usepackage{amsmath}
\usepackage{amssymb}
\usepackage{booktabs}
\usepackage[accsupp]{axessibility}

\usepackage[pagebackref=true,breaklinks=true,letterpaper=true,colorlinks,bookmarks=false]{hyperref}

\usepackage{iccv}             
\iccvfinalcopy
\def\iccvPaperID{6385} 

\ificcvfinal\fi

\input{macrosjvg}
\begin{document}
\title{Objects do not disappear: \\ Video object detection by single-frame object location anticipation}

\author{Xin Liu$^1$\ \ Fatemeh Karimi Nejadasl$^2$ \ \  Jan C. van Gemert$^1$\ \ Olaf Booij$^1$ \ \  Silvia L. Pintea$^{1}$\\ 
{\normalsize Computer Vision Lab, Delft University of Technology$^1$}\\
{\normalsize Institute for Biodiversity and Ecosystem Dynamics, University of Amsterdam$^2$}
}
\maketitle
\ificcvfinal\thispagestyle{empty}\fi

\input{Paper/abstract}
\input{Paper/intro2}
\input{Paper/related}

\input{Paper/method}
\input{Paper/experiments}
\input{Paper/conclusion}

{\small
\bibliographystyle{ieee_fullname}
\bibliography{egbib}
}

\newpage
\input{Paper/supplementary}
\end{document}

%% file: macrosjvg.tex
\usepackage[dvipsnames]{xcolor}

\newcommand{\rthree}[1]{{\bf \color{Maroon} R[g3ek]}}
\newcommand{\rtwo}[1]{{\bf \color{Plum} R[9Hbj]}}
\newcommand{\rone}[1]{{\bf \color{PineGreen} R[khHr]}}

\newcommand{\Eq}[1]{Eq.~(\ref{eq:#1})}
\newcommand{\eq}[1]{\Eq{#1}}
\newcommand{\fig}[1]{Fig.~\ref{fig:#1}}
\newcommand{\tab}[1]{Tab.~\ref{tab:#1}}

\usepackage{pifont}
\newcommand{\cmark}{\ding{51}}%
\usepackage{arydshln}

\usepackage{graphicx}
\usepackage{amsmath}
\usepackage{amssymb}
\usepackage{multirow}
\usepackage{booktabs}
\usepackage{comment}
\usepackage{siunitx}
\usepackage{multicol}
\usepackage{float}

\usepackage{soul}


%% file: Paper/abstract.tex
\begin{abstract}
Objects in videos are typically characterized by continuous smooth motion. 
We exploit continuous smooth motion in three ways. 
1) Improved accuracy by using object motion as an additional source of supervision, which we obtain by anticipating object locations from a static keyframe. 
2) Improved efficiency by only doing the expensive feature computations on a small subset of all frames. 
Because neighboring video frames are often redundant, we only compute features for a single static keyframe and predict object locations in subsequent frames. 
3) Reduced annotation cost, where we only annotate the keyframe and use smooth pseudo-motion between keyframes. 
We demonstrate computational efficiency, annotation efficiency, and improved mean average precision compared to the state-of-the-art on four datasets: ImageNet VID, EPIC KITCHENS-55, YouTube-BoundingBoxes and Waymo Open dataset. 
Our source code is available at {\small\href{https://github.com/L-KID/Video-object-detection-by-location-anticipation}{https://github.com/L-KID/Video-object-detection-by-location-anticipation}}.
\end{abstract}

%% file: Paper/intro2.tex
\section{Introduction}
Humans assume object permanence: blink your eyes, and the world is still there.  Similarly, video frames are redundant, and missing some frames when watching a movie does not drastically change the scene.
Actually, for the parts that \emph{did} change, if these parts changed coherently, they might hint at a sense of objectness, as hypothesized by the Gestalt law of common fate. 

\begin{figure}[t]
    \centering
    \includegraphics[width=.9\linewidth]{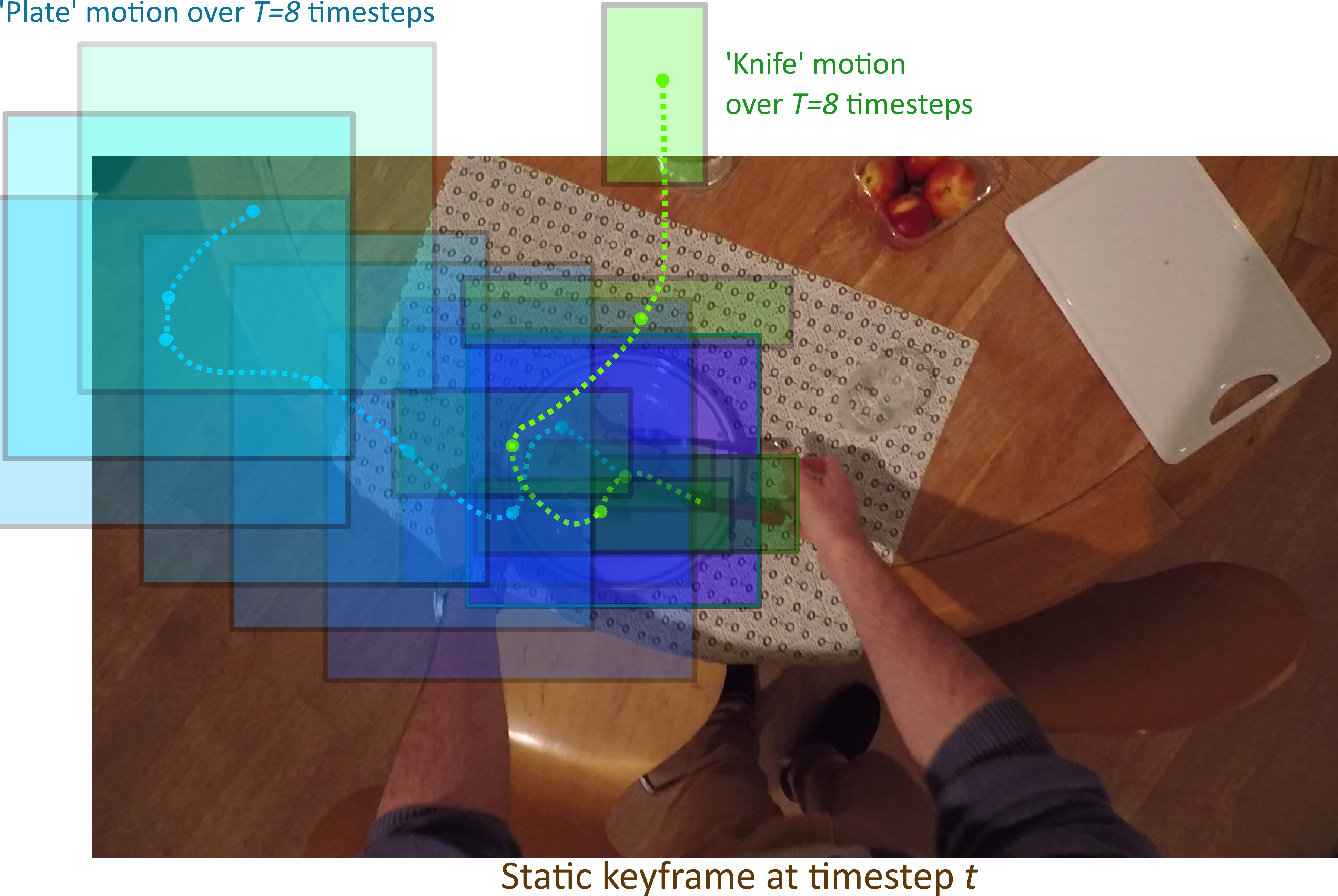}
    \caption{\small Anticipating future object locations from a static keyframe is \emph{efficient}.
        We only do the expensive feature extraction on a small subset of keyframes, while still accessing bounding-box locations for all video frames.
        Moreover, exploiting motion cues as additional supervision \emph{improves} object detection.
        By sampling a static keyframe at time $t$ and anticipating the object locations over the next $T$ timesteps, we incorporate temporal consistency and smoothness of object motion.
        }
    \label{fig:idea}
\end{figure}

As illustrated in \fig{idea}, here we explore these observations in the context of video object detection in three ways: 
1) Improved accuracy by exploiting an additional source of supervision: the coherent motion from the law of common fate; by predicting object motion from a static image. 
2) Improved efficiency by exploiting redundancy to reduce computational cost by only processing sampled keyframes and predict object motion for the missing frames in-between. 3) Reduced annotation cost by only annotating sampled keyframes. 
Thus, we improve accuracy and save computation and annotation time by simply skipping the feature computation and/or the annotation for a large majority of the frames. 

We make the following contributions:
(i) A video object detection method that samples static keyframes and predicts object motion for unseen future frames.
(ii) Computational efficiency, as our method only extracts features for sampled static keyframes;
(iii) Data efficiency, as we use sparse annotations only at the sampled keyframes, hallucinating motion in-between these sparse annotations.
(iv) Our extensive experimental results on ImageNet-VID \cite{ILSVRC15}, EPIC KITCHENS-55 \cite{damen2018scaling}, YouTube-BoundingBoxes \cite{real2017youtube} and Waymo Open dataset \cite{Ettinger_2021_ICCV} show that our approach improves accuracy over the state-of-the-art methods, while being faster at both training and inference time.

%% file: Paper/related.tex
\section{Related work}

\medskip\noindent\textbf{Video object detection.}
Several methods do temporal modeling by post-processing still-image object detectors such as post-processing  Faster-RCNN~\cite{han2016seq,kang2016object} or post-processing Mask RCNN~\cite{bertasius2020classifying,lin2020video}.
Alternatives include recurrent blocks \cite{lu2017online,tripathi2016context} or optical flow~\cite{zhu2017flow,zhu2018towards}. This can be further extended with instance and pixel-level calibrations over time \cite{wang2018fully}, or using a space-time lattice \cite{chen2018optimizing}.
More recently, the field has advanced by aggregating temporal information: either by defining detection correlations as a graph in SELSA \cite{wu2019sequence}, or by using global and local temporal pooling in MEGA \cite{chen2020memory}.
MEGA is further extended by considering all pairwise frames in TF-Blender \cite{cui2021tf}, while HVRNet \cite{han2020mining} integrates inter-video and intra-video object relations.
IFF-Net\cite{jin2022feature} uses a feature flow estimating module to indicate the feature displacement.
LWDN \cite{jiang2019deformable} does not simply aggregate features, but aligns the features between keyframes by adopting a memory mechanism.
In contrast, we avoid computationally demanding optical flow and recurrent blocks, and
we do not aggregate neighboring frames, or use temporal heuristics to post-process still-image detections.  
Instead, we anticipate future object locations over time from a single static keyframe, placing temporal prediction at the heart of our model.

\medskip\noindent\textbf{Single image motion prediction versus tracking.} Object tracking aims to predict object location in a video. 
A recent overview of object tracking is given in~\cite{ciaparrone2020deep}.  
In classical methods, the bounding box around the object to track is given a priori~\cite{smeulders2013visual}, but can also be estimated by an object detector \cite{bergmann2019tracking,feichtenhofer2017detect,held2016learning,sun2020transtrack}.
The object location in the next frame can be estimated with siamese networks~\cite{bergmann2019tracking}, or object detections at every frame are linked into tracks by motion regression~\cite{feichtenhofer2017detect}.
Our method is inspired by tracking, yet we are different because we predict the object location in future frames from a single static input image, without actually using the image features of the future frames.

\medskip\noindent\textbf{Anticipating motion from a static single image.}
A single static frame is rich enough to allow predicting of future appearance~\cite{hsieh2018learning,liu2021deep,narasimhan2022strumming,vukotic2017one}, actions~\cite{abu2018will,girdhar2021anticipative,mahmud2017joint,Singh2017ActionLocalisationandPrediction}, and motion~\cite{gao2018im2flow,pintea2014motion,walker2016uncertain}.
Such motion predictions, in turn, can then be used for predicting pedestrians' behaviors \cite{rasouli2019pie,zhang2020stinet} or other road agents~\cite{gu2021densetnt}. Moreover, motion prediction can be used as a self-supervision cue~\cite{feichtenhofer2021large,han2019video,han2020memory,li2021motion, wei2018learning} to improve feature learning. 
Inspired by these works, we use location anticipation as an additional source of supervision to improve the accuracy of the still-image detector, while we also exploit it for efficiency as it allows us to only compute features on a subset of still-images, avoiding expensive feature computations on all frames.

\medskip\noindent\textbf{Efficiency in video.}
Because videos typically sample several frames per second (FPS) it is important to have efficient video analysis methods.
Successful prior work proposes network architecture adaptations to reduce computations  \cite{feichtenhofer2020x3d,kondratyuk2021movinets,lin2019tsm,liuCVPR21noFrameBehind,pintea2016making,zolfaghari2018eco}.
For video object detection, it is efficient to adapt the detector online \cite{sharma2013efficient} or to transfer an image detector to video \cite{kalogeiton2016analysing}.
Alternatively, computations can be reduced by focusing only on specific video regions \cite{jiang2020learning,sun2022efficient, xu2020centernet}.
Similarly, we also focus on efficiency in video object detection. 
We are efficient by predicting object trajectories a few frames ahead and therefore saving computations by not processing those frames.

\medskip\noindent\textbf{Sparse annotations in video.}
Training a model on sparse video annotations can be done by iteratively updating the model in a boosting fashion \cite{all2011flowboost}, or propagating groups of pixels through time \cite{jain2014supervoxel}.
More recent work generates dense object masks in videos from sparse bounding boxes \cite{xu2020video}.
Rather than focusing on improving the model accuracy on sparsely annotated videos, 
prior work has analyzed what is the accuracy vs. annotation effort trade-off \cite{mettes2016spot}.
Specifically, for object locations in video, annotations can be generated through a combination of tracking, frame selection and active learning 
\cite{dai2021video,ince2021semi,kuznetsova2021efficient,vondrick2011video}.
Inspired by these works we also explore a variant of our model that allows sparse annotations. 
We require bounding-box annotations on the static keyframes, and we experimentally show that we can hallucinate the motion between keyframes, removing the need to annotate all frames.

%% file: Paper/method.tex
\begin{figure*}
    \centering
    \includegraphics[width=\linewidth]{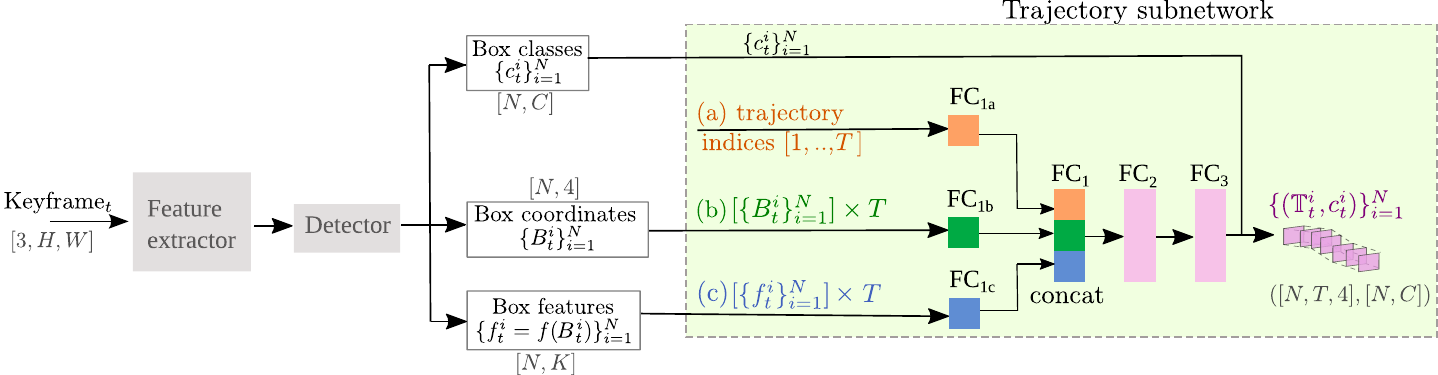}
    \caption{ \small
        \textbf{Overview:} The network starts from a single sampled video keyframe at timestep $t$.
        A feature extractor backbone is followed by an object detector.
        The object detector outputs for the keyframe at time $t$ a set of $N$ bounding boxes $\{B_t^i\}_{i=1}^N$, together with their associated class probabilities $\{c_t^i\}_{i=1}^N$ and keyframe features extracted from each box area $\{f_t^i\}_{i=1}^N$.
        Next, our trajectory subnetwork (highlighted in green) takes as input: 
        (a) a batch of trajectory indices $[1,..,T]$, where T is the trajectory length; (b) the keyframe bounding boxes repeated $T$ times and batched, $[\{B_t^i\}_{i=1}^N]{\times}T$; and 
        (c) the box features also repeated $T$ times $[\{f_t^i\}_{i=1}^N]{\times}T$.
        These are projected through linear layers of equal sizes (FC$_{1a}$, FC$_{1b}$, FC$_{1c}$) and the output is concatenated and passed through two additional linear layers.
        The trajectory network predicts a set of $N$ trajectories $\{\mathbb{T}_t^i\}_{i=1}^N$ of length $T$ and their associated classes $\{c_t^i\}_{i=1}^N$.
    }
    \label{fig:overview}
\end{figure*}

\section{Anticipating object locations}
\noindent\textbf{Object detection backbone.} 
We illustrate our method in \fig{overview}. 
We start from a standard static image object detection backbone. 
We input static frames uniformly sampled from the video with a time step $T$; we call such static frames ``keyframes''.
For each keyframe at time~$t$, the object detector gives a set of $N$ proposal bounding boxes: $B_t{=}\{{B}^i_t\}_{i=1}^N$ where ${B}^i_t{=}(x^i_t,y^i_t,w^i_t,h^i_t)$, and $(x^i_t,y^i_t)$ are the top-left corner and $(w^i_t,h^i_t)$ are the width and height of the bounding box.
Each $B^i_t$ corresponds to a possible object and has an associated class $c^i_t$: $\{({B}^i_t, c^i_t)\}_{i=1}^N$.

\smallskip\noindent\textbf{Trajectory subnetwork.}
Starting from the keyframe detection bounding boxes $B_t$, we anticipate object trajectories -- defined as the future bounding box locations of an object over the following $T$ frames.
The trajectory subnetwork is highlighted in green in \fig{overview}.
For each keyframe indexed by $t$, we define a batch of length $T$ and we input into the trajectory network three types of inputs:
(a) a vector of time indices of the future trajectories relative to the keyframe index $t$ batched from $[1, .., T]$; 
(b) the set of bounding boxes detected at the keyframe $B_t{=}\{{B}^i_t\}_{i=1}^N$ repeated over all $T$ time steps: $[\{{B}^i_t\}_{i=1}^N]{\times}T$; and
(c) the static keyframe featuremaps extracted via the network mapping $f(\cdot)$ from the the bounding boxes $\{f^i_t{\mid} f^i_t{=}f({B}^i_t)\}_{i=1}^N$, also broadcast for each of the $T$ time steps: $[\{{f}^i_t\}_{i=1}^N]{\times}T$.
Note that by inputting into the trajectory subnetwork the future trajectory-frame indices in (a), we add temporal ordering information: each future box prediction has an associated frame index.
We map all three inputs: boxes, features and time indices, through fully connected (FC) layers of equal sizes. 
We concatenate the output features, and pass them through two additional fully connected layers.

The output of the trajectory subnetwork is a list of trajectories, one trajectory for each of the N detected objects.
Each trajectory starts at the keyframe indexed by $t$ and extends up to frame $t{+}T$.
Concretely, our trajectory predictions are: $\{ \mathbb{T}^i_t{\mid}\mathbb{T}^i_t{=}(B_t^i, {B}^i_{t+1}, .., {B}^i_{t+T}) \}_{i=1}^N$, where we also add the keyframe bounding box $B_t$ to the trajectory. 

\subsection{Associating trajectories to ground truth}
To optimize the trajectories we need an associated object class for each trajectory.
Each object trajectory $\mathbb{T}^i_t$, indexed by $i$, starts with a keyframe bounding box $B^i_t$, which has a corresponding object class $c^i_t$. 
Each trajectory corresponds to one object, thus we let each trajectory inherit the class of its starting keyframe bounding box, yielding: $\{(\mathbb{T}^i_t, c^i_t)\}_{i=1}^N$. 

Moreover, we also need to associate ground truth boxes $B^{*}$ with all predicted boxes along each object trajectory $\mathbb{T}^i_t{=}(B_t^i, B^i_{t+1}, .., B^i_{t+T})$.
Following the standard procedure \cite{he2017mask, ren2015faster} we rank the predicted boxes $B_{t+l}, l{\in}\{0, .., T\}$ based on their overlap with the ground truth boxes $B_{t+l}^*$ at each frame $t{+}l$.
We associate each of the $N$ predicted boxes with the best matching IoU score of the ground truth box. 

\subsection{Trajectory loss}
\noindent\textbf{Bag of boxes loss.}
We want to optimize for each object indexed by $i$ its associated trajectory, starting at keyframe $t$: $\mathbb{T}^i_{t}$. 
For readability, we ignore the index $i$ from here on. 
The standard loss $L_{\text{bag}}$ for performing box regression, considers the trajectory boxes as an unordered bag and computes a smooth $L_1$ loss $\mathcal{L}_1(\cdot)$ \cite{girshick2015fast,ren2015faster}, for each predicted box $B_{t+l}$, to its associated ground truth box $B^*_{t+l}$: 
\begin{align}
    L_{\text{bag}}(B^*, \{B_t, .., B_{t+T}\}) = \sum_{l=0}^{T} \mathcal{L}_1({B}^*_{t+l} - {B}_{t+l}).
    \label{eq:loss}
\end{align}

\begin{figure}
    \centering
    \includegraphics[width=1.\linewidth]{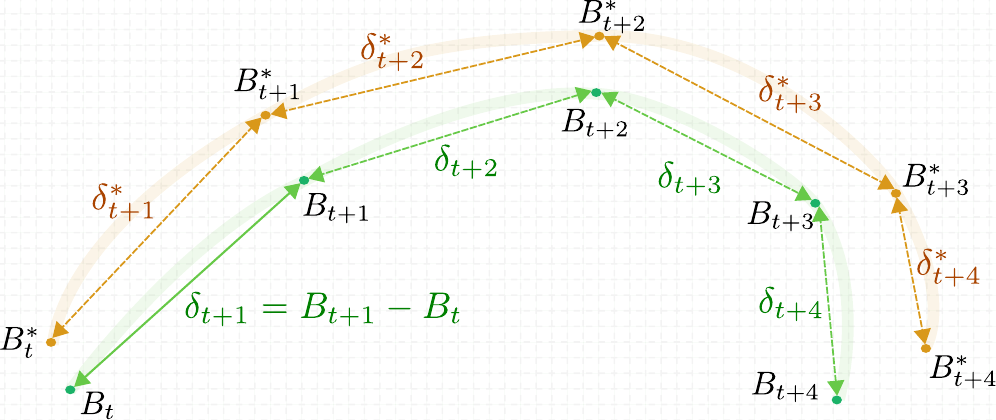}
    \caption{ \small
        Object trajectories are piecewise continuous: \ie an object cannot disappear between two neighboring frames $t$ and $t{+}1$.
        We incorporate this continuity by noting that the loss between a box $B_{t+l}$ along the trajectory and its associated ground truth $B^*_{t+l}$ is defined as the sum of pairwise offsets of neighboring boxes, $\delta_{t+l}$, starting from the keyframe box $B_t$.
        The orange line is the true trajectory and the green line is the predicted trajectory. 
        (Here, for simplicity we discard the width and height of the bounding boxes and only show $(x,y)$ coordinates and depict $l{\in}\{1,..,4\}$.)
    }
    \label{fig:loss}
\end{figure}

\noindent\textbf{Trajectory cumulative loss.}
The downside of \eq{loss} is that it treats each prediction $B_{t+l}$ as if it were independent of its neighboring predictions along the trajectory, $B_{t+l-1}$ and $B_{t+l+1}$.
Therefore, there is no temporal ordering enforced in the $L_\text{bag}$ loss. 
Not enforcing the ordering of the predictions along the trajectory could lead to discontinuous trajectories. 
We want to enforce smoothness in the predictions over time: objects cannot disappear or appear at random locations, between neighboring frames. 
Specifically, the trajectory is piecewise continuous: for an object to move from a location $B_{t+1}$ to a location $B_{t+4}$ it has to travel through the intermediate locations $B_{t+2}$ and $B_{t+3}$, as illustrated in \fig{loss}. 

To add this insight, we define a loss that constrains the pairwise offsets along the trajectory from frame $t$ up to every frame $t{+}l$: $\delta_{t+k}{=}(B_{t+k}{-}B_{t+k-1})$, $k{\in}\{1, .., l\}$, to add up to the offset from the ground truth at the frame $t{+}l$ to the keyframe prediction $(B^*_{t+l}{-}B_t)$.
Concretely:
\begin{align}
    \label{eq:lsum}
    L_{\sum}(B^*, \overleftrightarrow{\mathbb{T}_t}) =
        & \sum_{l=0}^T {\mathcal{L}_1}\left( \left( {B}^*_{t+l} - B_t \right) - \sum_{k=1}^{l} \delta_{t+k} \right),
\end{align}
where we redefine the trajectory to predict $\delta_{t+l}$ values describing pairwise offsets instead of bounding boxes: $\overleftrightarrow{\mathbb{T}_t}=({\delta}_{t+1}, .., {\delta}_{t+T})$.
Additionally, we ignore the calculation of the coordinates, if the ground truth bounding boxes ${B}^*_{t+l}$ are not valid in \eq{lsum}.
Note, that if we would predict bounding boxes $B_{t+l}$ in the trajectory network, instead of offsets between pairs of bounding boxes $\delta_{t+l}$, \eq{lsum} would reduce to \eq{loss} and the temporal ordering would not be enforced (See the derivation in the supplementary material).

In \eq{lsum} the inner loop over pairwise offsets $\delta_{t+k}$ accumulates the errors in the predictions over time from frame $t{+}1$ to frame $t{+}l$. 
To make sure the errors do not accumulate along the trajectory, and the change from frame to frame is smooth, we add another loss $L_{\text{bag}(\delta)}$ constraining the pairwise offsets $\delta_{t+l}$ at every timestep $t{+}l$ to map back to ground truth offsets: $\delta^*_{t+l}{=}({B}^*_{t+l}{-}{B}^*_{t+l-1})$:
\begin{align}
    \label{eq:ldelta}
    L_{\text{bag}(\delta)}(B^*, \overleftrightarrow{\mathbb{T}_t}) = \sum_{l=1}^{T} \mathcal{L}_1 \left({\delta}^*_{t+l} - \delta_{t+l} \right).
\end{align}

Our final loss $L_\text{traj}$ is then a combination of the two losses where the cumulative trajectory loss $L_{\sum}$ enforces piecewise continuity in the predictions and the bag of offsets loss $L_{\text{bag}(\delta)}$ discourages errors from accumulating along the trajectory and makes the trajectory smooth:
\begin{equation}
    L_\text{traj} = L_{\sum} + L_{\text{bag}(\delta)}.  
    \label{eq:traj-loss}
\end{equation}
We investigate the effect of each loss term in the experimental section, together with the effect of predicting offsets $\delta_{t+l}$ instead of box coordinates $B_{t+l}$ in the trajectory network.

\medskip\noindent\textbf{Sparse annotation loss.}
When there are no annotations in-between keyframes then we cannot optimize the anticipated trajectory. 
Since the task in the sparse annotation cases is object detection on annotated keyframes, we can hypothesize that the precise true location of an object along a trajectory ${B}^*_{t+l}$, is not essential, as long as the trajectory is piecewise continuous and smooth, and the starting and ending points of the trajectory are known. 
Therefore, we can rewrite our losses in \eq{traj-loss} to a sparsely-annotated variant $L_\text{traj}^{\text{(sa)}}$ by changing the way in which we define the box-supervision.  
Explicitly, we replace the set of ground truth boxes $\{B_t^*, B_{t+1}^*, .. B_{t+T}^*\}$ with a pseudo box-trajectory.
The pseudo-box trajectory is defined by a continuous function, $r_t(\cdot)$ describing the trajectory at every timestep as: $\mathbb{T}_{r_t}{=}(B^*_t,r_{t+1}(B^*_{t}),.. r_{t+T}(B^*_{t}))$, relative to the true keyframe location $B^*_t$. 
Because the next keyframe is the last trajectory location, we also constrain the pseudo trajectory to match the true bounding box at the end of the trajectory: $r_{t+T}(B^*_{t}){=}B^*_{t+T}$.

In practice, we choose $r_t(\cdot)$ to be either linearly interpolated box annotations, or a parabola as it is continuous and does not assume linear object trajectories.
Here, we only need bounding-box annotations every $T$ frames, for correcting the starting- and end-point of our predicted trajectory. 
Our sparsely-annotated loss $L_\text{traj}^\text{(sa)}$ variants are useful when the dataset only has sparse annotations available.

%% file: Paper/experiments.tex
\section{Experiments}
\noindent\textbf{Datasets and evaluation setup.}
We test our hypotheses on a fully controlled MovingDigits dataset. 
We ablate model choices on a subset of ImageNet VID~\cite{ILSVRC15}. 
We show the practical benefits of our approach on the full ImageNet VID~\cite{ILSVRC15} and on EPIC KITCHENS-55~\cite{damen2018scaling}. 
As a realistic sparsely annotated scenario, we evaluate on YouTube-BoundingBoxes ~\cite{real2017youtube} which has approximately 1 keyframe annotation per second, and the Waymo Open dataset \cite{Ettinger_2021_ICCV}.
For the ImageNet VID experiments, we train our model on a combination of ImageNet VID and DET datasets as is common practice in \cite{cui2021tf, han2020mining, wu2019sequence, zhu2017flow}. 
To quantify detection accuracy we adopt the common approach ~\cite{chen2020memory,cui2021tf,wang2018fully,wu2019sequence} by computing mean Average Precision (mAP), where a detection is correct if its Intersection over Union (IoU) with the ground truth is sufficiently large. 
Note that although our method samples keyframes during training, we evaluate on all frames at test time, unless stated otherwise.

\smallskip\noindent\textbf{Implementation details.}
We test either using a Faster RCNN \cite{ren2015faster} detector or a Deformable DETR \cite{zhu2020deformable} with an ImageNet pre-trained ResNet \cite{he2016ResNet101} or SwinB \cite{liu2021swin} base as the object detection backbone. 
Our trajectory prediction sub-network contains three fully-connected layers with 1024 dimensions for the middle layer, see \fig{overview}. 
We train our network for 120k iterations on ImageNet VID and 173k iterations on Epic Kitchens with the SGD optimizer, on 4 GPUs.
For ImageNet VID, the initial learning rate is \num{e-3} and is divided by 10 at 80K iterations.
For Epic Kitchens, the initial learning rate is \num{5e-4} and is divided by 10 at 120K iterations.
For YouTube-BoundingBoxes, the initial learning rate is \num{5e-4} and is divided by 10 at 100k iterations.

\subsection{Hypothesis testing}
\begin{figure}
    \centering
    \begin{tabular}{ccc}
    \small{Input frame} & \small{GT Bbox} & \small{Predicted Bbox} \\
    \includegraphics[width=0.3\linewidth]{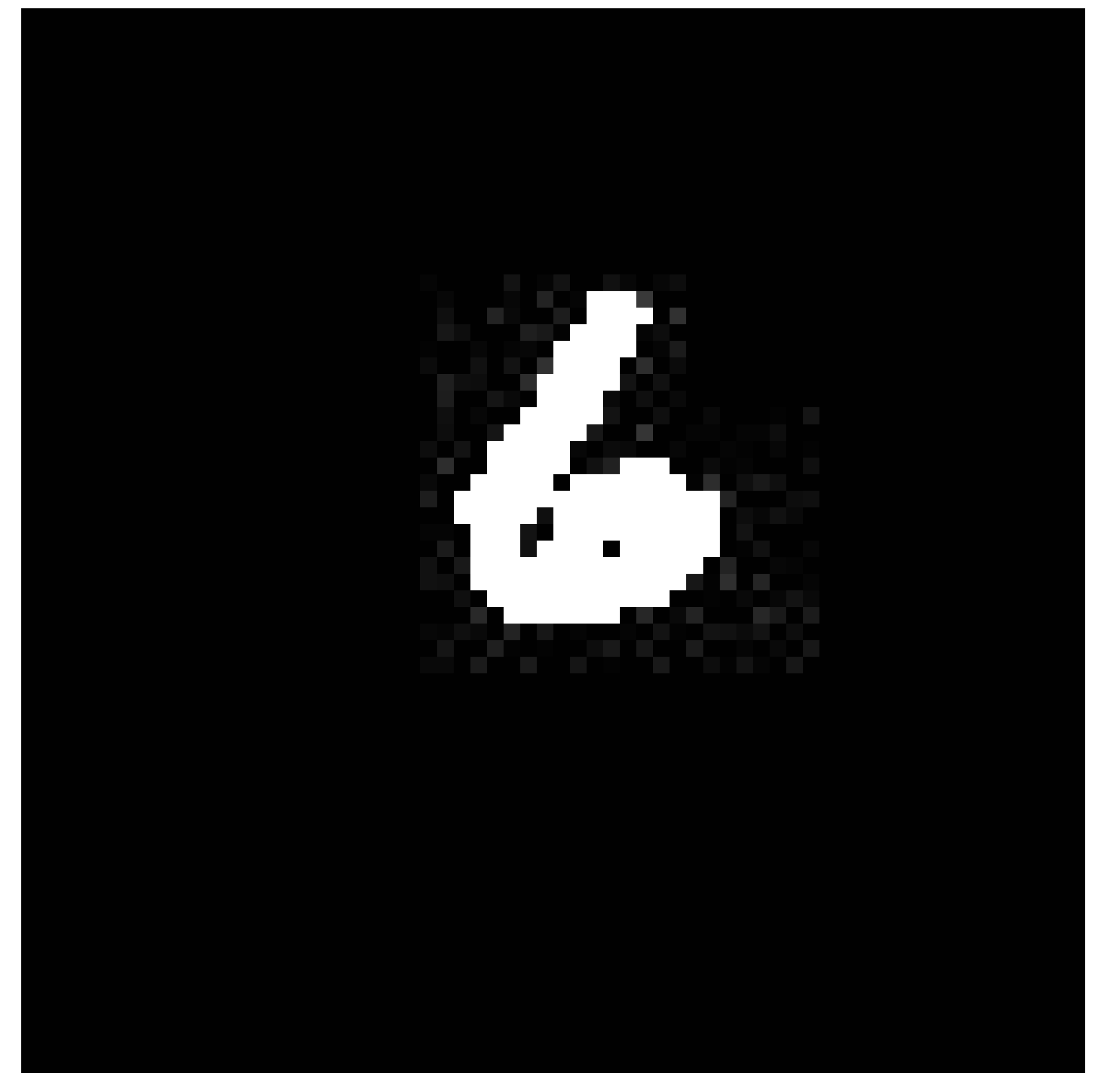} & 
    \includegraphics[width=0.3\linewidth]{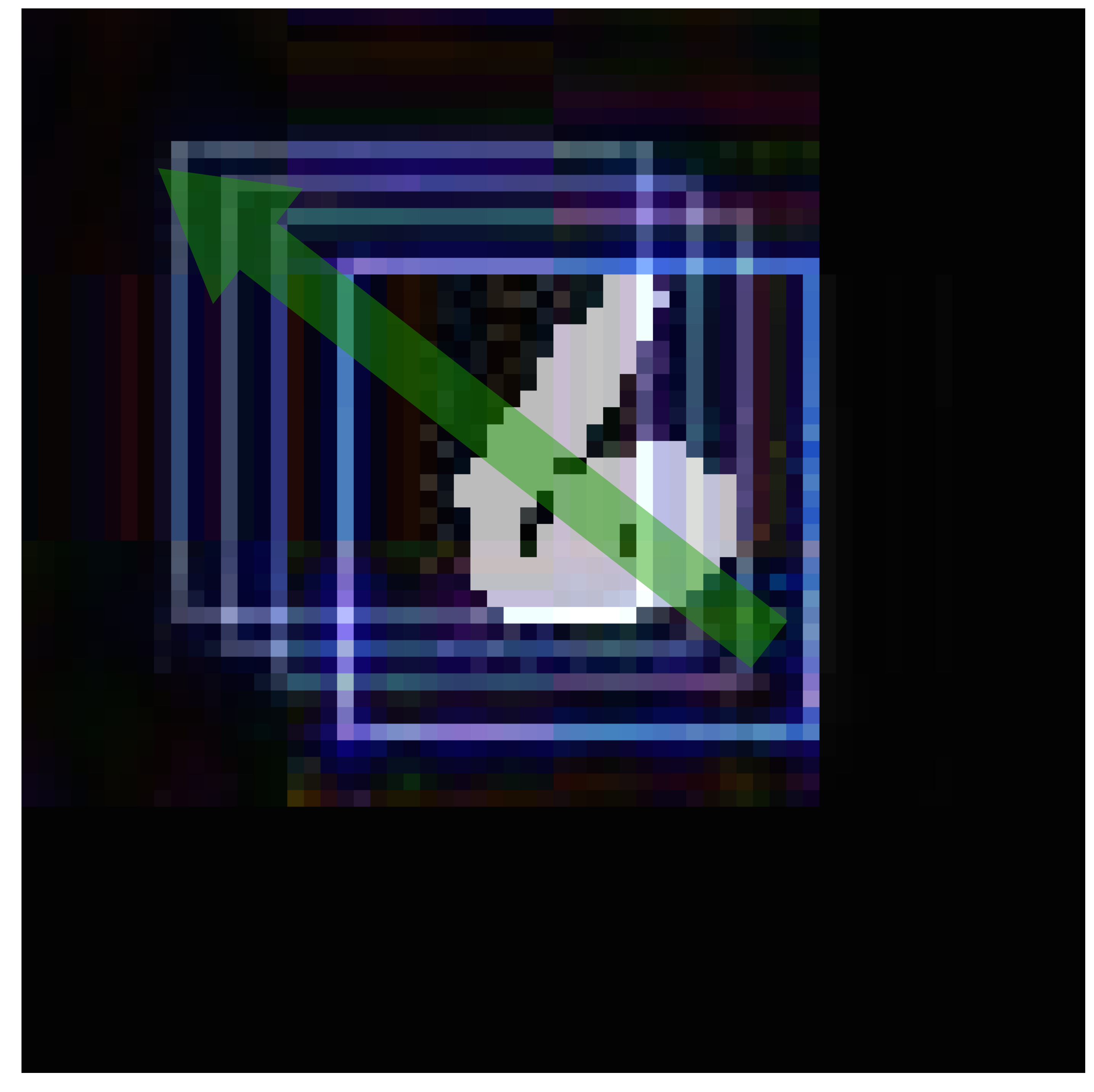} &
    \includegraphics[width=0.3\linewidth]{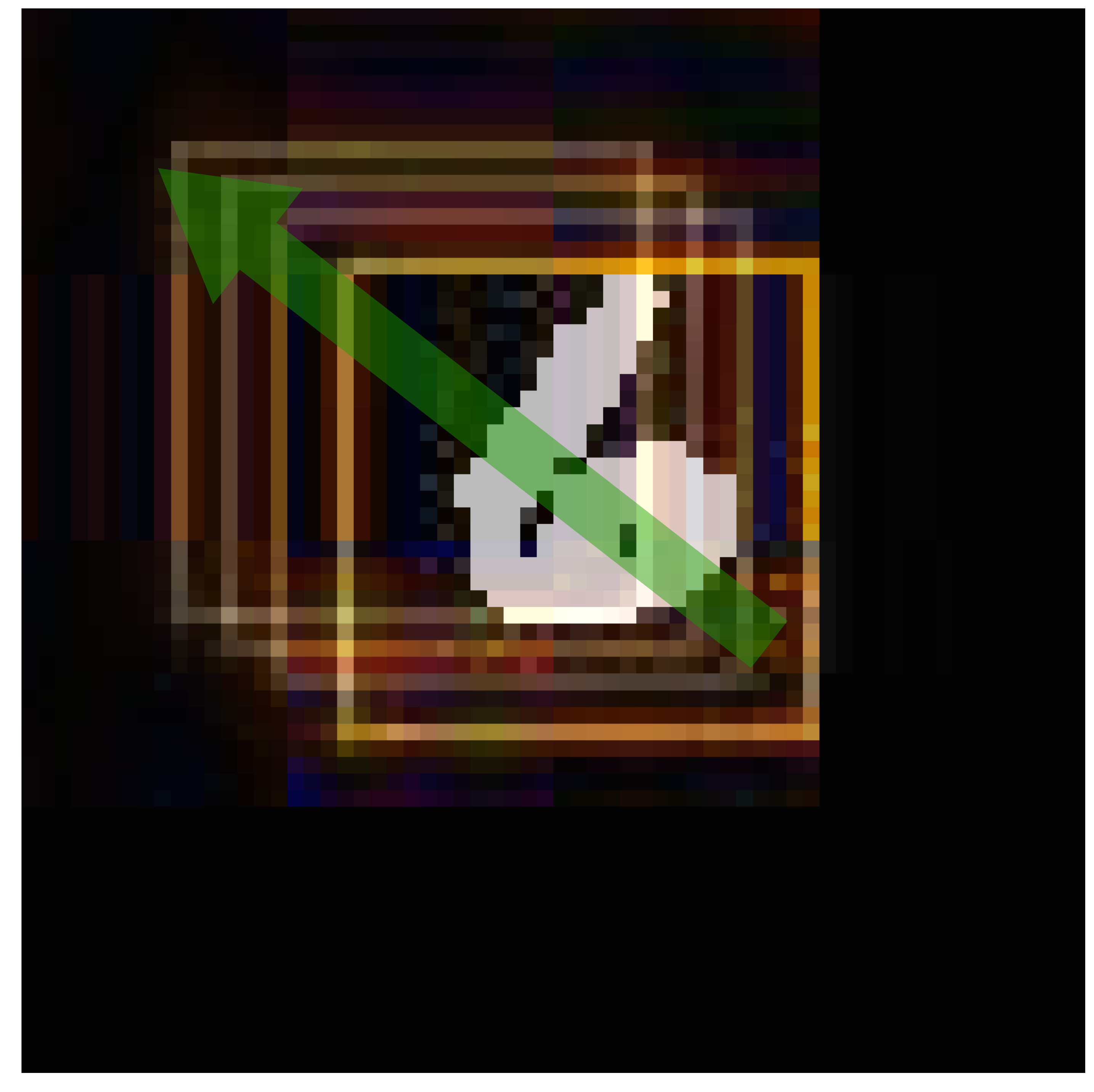} \\
    \includegraphics[width=0.3\linewidth]{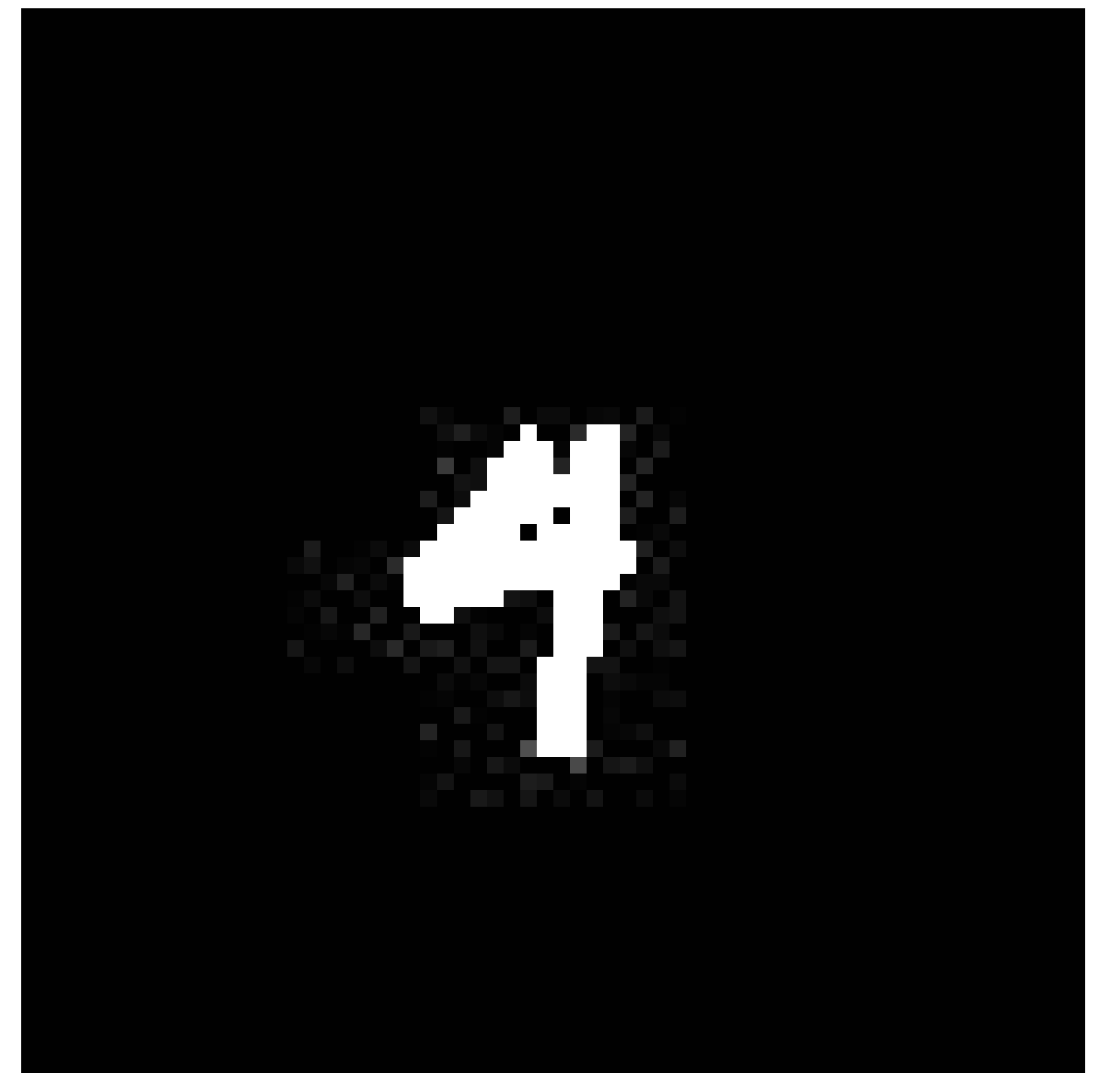} &
    \includegraphics[width=0.3\linewidth]{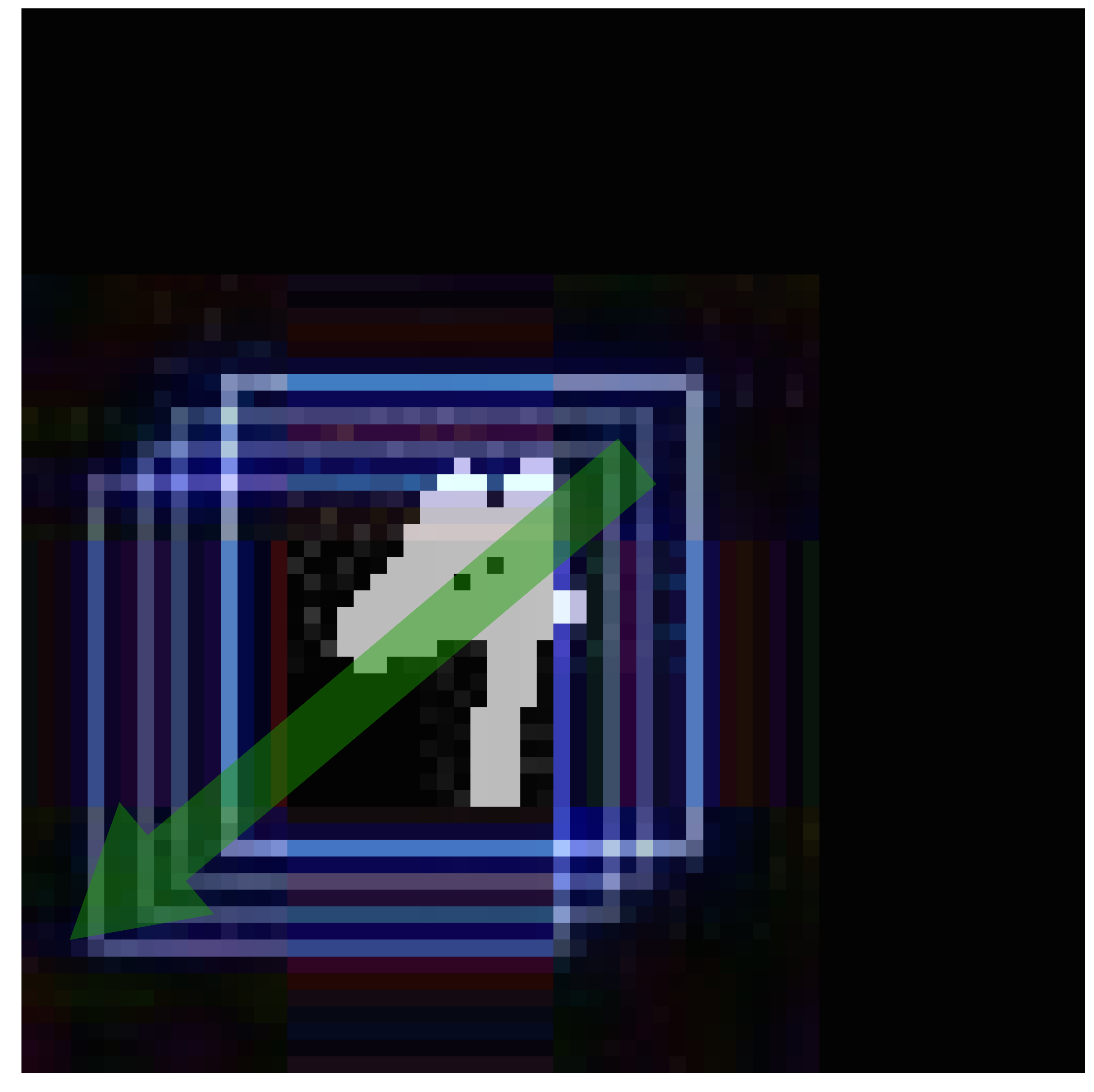} &
    \includegraphics[width=0.3\linewidth]{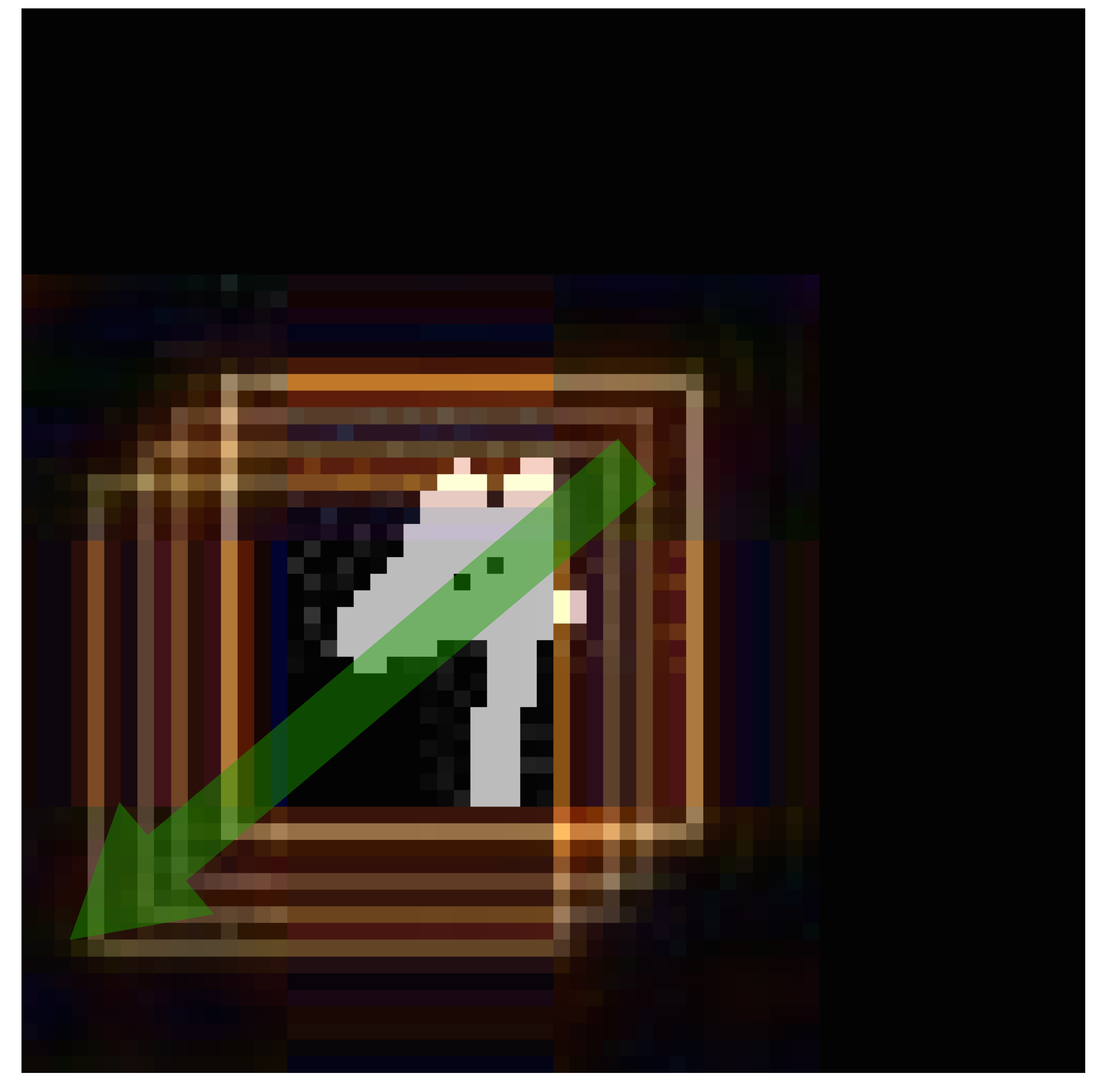} \\
    $t$ & $t \colon t{+}3$ & $t \colon t{+}3$ \\
    \end{tabular}
    \caption{ \small
    \textbf{[H1]: Trajectory anticipation on MovingDigits.} We show a single static input frame of digits 6 and 4 and their time-accumulated ground truth and predicted bounding boxes in timesteps $t \colon t{+}3$. Each digit class has an associated linear motion (green arrow). The match of our predicted bounding boxes and the ground truth bounding boxes shows that our model can predict trajectories from a static frame. 
    }
    \label{fig:h1}
\end{figure}

We test our hypotheses by creating our own fully controlled dataset, MovingDigits, where we pair each of the 10 MNIST digit classes with a unique, linear motion of 2 px per frame, see \fig{h1}.
Each video has 32 frames with a frame size of 64$\times$64 px.
We created 200 videos for training and 80 videos for testing, with an equal number of videos per class. We use a trajectory length of $T{=}8$, train for  1.25k iterations and use a ResNet-18 feature extractor and a Faster RCNN detector.
We cannot use the existing MovingMNIST \cite{srivastava2015unsupervised} because it does not provide detection bounding boxes, and it does not contain motion-appearance correlations.

\medskip\noindent\textbf{[H1]: Can the model anticipate motion trajectories?}
To verify if our model can anticipate trajectories from a single static input frame $t$, we train on our MovingDigits, where each digit has its own linear motion. 
We calculate the average IoU over the predicted trajectories for the test videos. 
The IoU is 0.95, which is near-perfect compared to the IoU of 0.79 for no motion anticipation. 
We show an example of the predicted bounding boxes (Bbox) by our method and the ground truth bounding boxes (GT Bbox) from time steps $t$ to $t{+}3$ in \fig{h1}.
Our model can successfully learn motion by anticipating trajectories from a static input frame. 

\medskip\noindent\textbf{[H2]: Anticipating improves static detection.} 
The motion cues in-between the static keyframes offer an additional source of supervision. 
Here, we investigate how the motion anticipation affects the static object detector, when we evaluate at test-time only on keyframes.
We consider four types of motion supervision for predicting box trajectories: (1) \textit{Ground truth motion}: trained on true bounding-box trajectories annotated at every frame; (2) \textit{Simulated smooth motion}:  bounding boxes move between keyframes according to a smooth parabola. (Details and examples are in the supplementary material); (3) \textit{Randomized positions}: there is motion, but it is not smooth, the boxes in-between the keyframes can occur at any random position in the image; (4) \textit{No motion}: without motion prediction, \ie, the static object detector baseline trained at every video frame. 

\tab{smoothandcontinuity} shows the keyframe mAP scores for IoU@ $[0.50{:}0.05{:}0.95]$. 
The \textit{No motion} static object detector is the baseline, which uses no motion.
The \textit{Randomized positions} as supervision is detrimental for object detection because the motion is random, and unpredictable. 
Interestingly, both \textit{Ground truth motion} and \textit{Simulated smooth motion} supervision improve the static keyframe detection. 
We speculate that the anticipation loss encourages detecting static regions that are most likely to move coherently, with consistent motion offsets: i.e. same direction.
Thus, for non-random motions, adding motion anticipation as additional supervision improves static object detection at keyframes, even without knowing the ground truth motion between the keyframes.

\begin{table}[]
    \centering
    \resizebox{0.75\linewidth}{!}{%
    \begin{tabular}{lc}
    \toprule
        Motion & Keyframe mAP (\%)  \\ \midrule  
         Randomized positions &  62.68\\
         No motion &  73.51\\
         Simulated smooth motion &  76.57\\
         Annotated motion & 79.31\\
    \bottomrule
    \end{tabular}}
    \vspace{1.5ex}
    \caption{ \small
    \textbf{[H2] Influence of motion anticipation on static object detection.} 
    Static keyframe detection mAP on MovingDigits for varying motion type supervision.
    For non-random motions, anticipating motion improves static object detection at keyframes.
    }
    \label{tab:smoothandcontinuity}
\end{table}

\subsection{Ablation of model components}
We run all ablation experiments on an ImageNet VID \cite{ILSVRC15} subset containing the classes `dog', `giant\_panda', and `hamster'. We use a ResNet-101 for feature extraction and Faster RCNN for detection.

\smallskip\noindent\textbf{[A1]: The effect of the trajectory loss.}
We evaluate  each term in our trajectory loss $L_\text{traj} = L_{\sum}{+}L_{\text{bag}(\delta)}$ in \eq{traj-loss}, compared to the standard loss $L_{\text{bag}}$ in \eq{loss}. 
The mAP scores for trajectory lengths $T{=}4$ and $T{=}8$ are in \tab{loss_comparison}.
The $L_\text{traj}$ loss has a higher mAP than the bag loss $L_{\text{bag}}$ for both trajectory lengths.
$L_\text{traj}$ enforces continuity and smoothness in the predictions over time thus leads to more precise predictions.
Furthermore, the improvement of $L_\text{traj}$ over $L_{\text{bag}}$ is larger for longer trajectories, e.g. for $T{=}4$ and $T{=}8$, $L_\text{traj}$ outperforms $L_{\text{bag}}$ by 1.66\% and 2.18\% respectively.
Our $L_\text{traj}$ loss is defined using offsets, without offsets the $L_{\sum}$ would be equivalent to $L_{\text{bag}}$. 
We observe that the effect of predicting offsets between neighboring boxes $\delta_{t+l}$, instead of bounding-box coordinates $B_{t+l}$ gives an improvement of 0.81\% and 0.89\%. 
When considering $L_{\text{bag}(\delta)}$ or $L_{\sum}$ individually, their mAP is lower or on par with $L_\text{traj}$.
This is because $L_{\text{bag}(\delta)}$ cannot enforce trajectory continuity, while $L_{\sum}$ cannot ensure that trajectories do not diverge by accumulating errors over time.  
Predicting offsets instead of box coordinates results in better accuracy.
\begin{table}
    \centering
    \resizebox{0.65\linewidth}{!}{%
    \begin{tabular}{l c c}
    \toprule
                                    & \multicolumn{2}{c}{Trajectory  length (mAP \%) }\\ \cmidrule(lr){2-3}
        Loss                        & T=4                & T=8  \\ \midrule
        \textbf{$L_{\text{bag}}$}   & 87.54 $\pm$ 0.16   & 83.97 $\pm$ 0.71  \\ \midrule
        $L_{\sum}$                  & 87.95 $\pm$ 0.27   & 84.09 $\pm$ 0.67 \\
        $L_{\text{bag}(\delta)}$    & 85.19 $\pm$ 0.66   & 79.73 $\pm$ 1.01 \\
        $L_\text{traj}$    &  \textbf{89.20 $\pm$ 0.21}  & 86.15 $\pm$ 0.75 \\
    \bottomrule
    \end{tabular}}
    \vspace{1.5ex}
    \caption{ \small
    \textbf{[A1]: Loss choice.}
    Compared to $L_{\text{bag}}$, the $L_{\text{bag}(\delta)}$ loss does worse, whereas $L_{\sum}$ performs on par. Their combination in  $L_\text{traj} = L_{\sum}{+}L_{\text{bag}(\delta)}$ outperforms $L_{\text{bag}}$.
    Predicting offsets instead of bounding-box coordinates in the trajectory network gives better results.
    These patterns are consistent over both trajectory lengths. 
    }
\label{tab:loss_comparison}
\end{table}

\medskip\noindent\textbf{[A2]: The effect of the sparse annotation loss.}
We test our model using the $L_\text{traj}^{\text{(sa)}}$ loss for sparse annotations.
To evaluate this in a controlled setting, we sub-sample keyframe annotations from the fully annotated ImageNet VID subset to mimic a sparse annotation scenario, and evaluate only on keyframes. 
We compare the ground truth motion with the result of anticipating linearly interpolated trajectories between keyframes.
\tab{a2-sa} shows that the $L_{\text{bag}(\delta)}^{\text{(sa)}}$ does not contribute much for the keyframe detection.
This is because $L_{\text{bag}(\delta)}^{\text{(sa)}}$ constrains the predicted offsets to match the pairwise offsets of the pseudo trajectory at every frame, which is not useful here since the motion is simulated.
The $L_{\sum}^{\text{(sa)}}$ performs on par with the $L_\text{traj}^{\text{(sa)}}$.
And the keyframe detection accuracy with the sparse annotation loss $L_\text{traj}^{\text{(sa)}}$ is close to that with the proposed fully-supervised loss $L_\text{traj}$.

\begin{table}
    \centering
    \resizebox{\linewidth}{!}{%
    \begin{tabular}{l c c c c}
    \toprule
        & \multicolumn{4}{c}{Keyframe mAP\%}\\ \cmidrule(lr){2-5}
        & \multirow{2}{*}{$L_\text{traj}$} & \multicolumn{3}{c}{Sparse annotation loss}\\ \cmidrule(lr){3-5}
                  &   & $L_{\sum}^{\text{(sa)}}$ &  $L_{\text{bag}(\delta)}^{\text{(sa)}}$ &  $L_\text{traj}^{\text{(sa)}}$ \\ \midrule           
        T=4     &  91.03 $\pm$ 0.19 & 90.35 $\pm$ 0.34 & 74.61 $\pm$ 0.72 & 90.76 $\pm$ 0.26 \\             
        T=8     &  87.94 $\pm$ 0.66 & 86.98 $\pm$ 0.73 & 68.76 $\pm$ 0.99 & 87.12 $\pm$ 0.69 \\
    \bottomrule
    \end{tabular}}
    \vspace{0.05ex}
    \caption{ \small
    \textbf{[A2]: Sparse annotation loss analysis.}
    We sub-sample keyframes of the fully annotated ImageNet VID subset to mimic the sparse annotations, and evaluate our sparse annotation loss on keyframe detection. 
    The $L_{\text{bag}(\delta)}^{\text{(sa)}}$ does not improve the detection accuracy, while $L_{\sum}^{\text{(sa)}}$ performs on par with the $L_\text{traj}^{\text{(sa)}}$. 
    The sparse annotation loss achieves a comparable accuracy on keyframe detection with our fully-supervised loss $L_\text{traj}$.
    }
\label{tab:a2-sa}
\end{table}

\begin{figure}
    \centering
    \includegraphics[width=.9\linewidth]{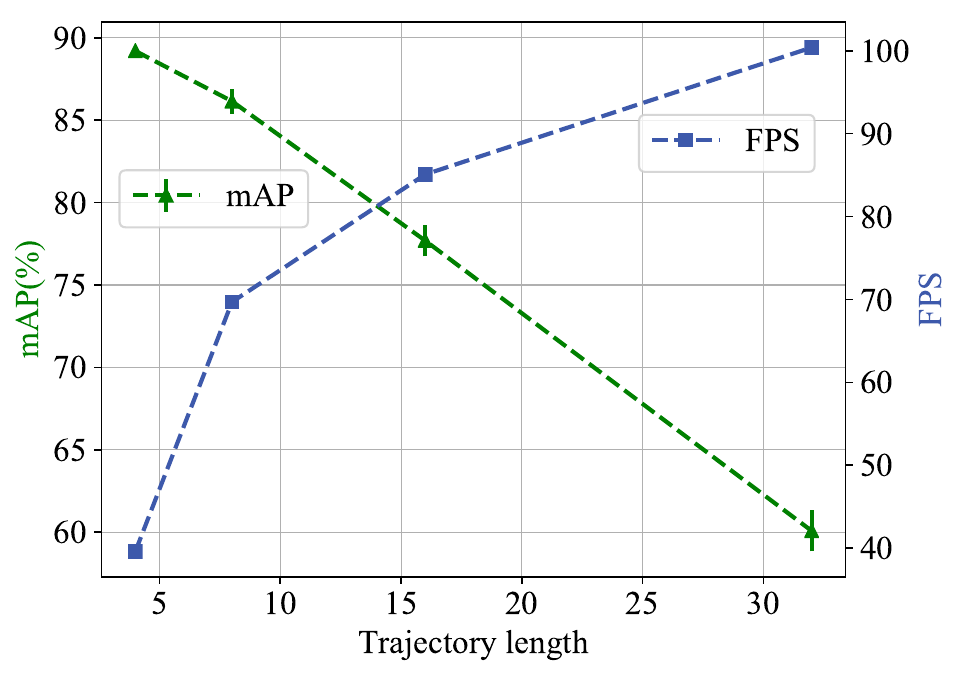}
    \caption{ \small
    \textbf{[A3] Inference speed vs. accuracy trade-off.}
    We can increase inference speed (FPS) by sampling fewer keyframes and predicting longer trajectories. Yet, by increasing trajectory length the mAP decreases. 
    The trajectory length can be selected according to the needed speed-accuracy trade-off.
    }
    \label{fig:speed-acc}
\end{figure}

\medskip\noindent\textbf{[A3]: Inference speed vs. accuracy trade-off.}
We can control the inference speed by sampling fewer keyframes, and thus predicting longer trajectories.
We analyse the speed vs. accuracy trade-off of our method on the subset of ImageNet VID.
\fig{speed-acc} shows we can reach an mAP of 89.2\% with a runtime of 39.6 FPS.
Moreover a noticeable drop in mAP occurs at trajectories longer than 10 frames.
However, when the trajectory is too long, the object motion may vary or the object may leave the frame or a new object may enter the frame. 
These changes result in the decrease of our model\rq s performance.
The trajectory length can be chosen according to the desired speed-accuracy trade-off.

\subsection{Comparison with state-of-the-art}
\noindent\textbf{[C1]: Experiments on ImageNet VID.}
We compare our method with state-of-the-art video object detection methods on ImageNet VID  in \tab{ImageNetVID}.
We mark methods without post-processing with \cmark.
Methods with post-processing add extra computational cost.
Our method does not use any post-processing.
We use a prediction trajectory of length $T{=}4$, as this already gives a considerable reduction in computation speed.
For our model, we test using a ResNet-101 \cite{he2016ResNet101} and a SwinB \cite{liu2021swin} backbone with either a Faster RCNN \cite{ren2015faster} or a Deformable DETR \cite{zhu2020deformable}. 

Among all methods using a ResNet-101 backbone, our method is the most accurate with a 87.2\% mAP, which has a 2.7\% improvement over the leading MEGA \cite{chen2020memory}.
We also report runtime (FPS) and train-time (hrs/epoch) to show that our method is fast and efficient.
We measure the efficiency using the code provided by the original papers, and where the code is not available we mark this with `-'. 
Note that for the training time this is a rough estimate, when considering the same settings (batch-size, GPU) for all methods. 
We tested the inference runtime speed on a single NVIDIA GTX 1080 Ti.
In terms of both training time and runtime our method is most efficient.
Our method has fast inference time with 39.6 FPS, which is $\approx 1.8\times$ faster than existing fast methods like LWND \cite{jiang2019deformable} and ST-Lattice \cite{chen2018optimizing}.
Our efficiency comes from predicting object locations for the next $T$ frames, while processing only sub-sampled keyframes.

Because our method is detector agnostic, we also compared our method with methods using a more advanced detector, Deformable DETR \cite{zhu2020deformable}, and a more advanced backbone SwinB \cite{liu2021swin}. 
In \tab{ImageNetVID} our method is on-par with PTSEFormer \cite{wang2022ptseformer} using the Deformable DETR detector and R101 backbone. 
And our method outperforms overall the state-of-the-art, when using a SwinB backbone.
\begin{table}[]
    \centering
    \resizebox{\linewidth}{!}{%
    \begin{tabular}{l@{\hskip -0.05in}c@{\hskip 0.in}c@{\hskip 0.1in}c@{\hskip 0.1in}c@{\hskip 0.1in}r@{\hskip 0.1in}r}
    \toprule
        Methods &  Backbone & No Post- & mAP (\%)  & Train-time & Runtime \\ 
         &   & proc. & & (hrs/epoch) & (FPS) \\ \midrule
         Faster-RCNN \cite{ren2015faster}& R101 & \cmark & 73.6 & 1.55 & 21.2\\
         LWND \cite{jiang2019deformable} & R101 & \cmark & 76.3 & - & 20.0\\
         FGFA \cite{zhu2017flow}& R101 &  & 78.4 & 6.59 & 5.0 \\
         THP \cite{zhu2018towards}& R101+DCN & \cmark & 78.6 & - & - \\
         ST-Lattice \cite{chen2018optimizing} & R101 &  & 79.6 & 1.40 & 20.0\\
         D\&T \cite{feichtenhofer2017detect}& R101 &  & 80.2 & 6.56 & 5.0\\
         MANet \cite{wang2018fully} & R101 &  & 80.3 & 6.88 & 4.9 \\
         STSN \cite{bertasius2018STSN}& R101+DCN &  & 80.4 & - & -\\
         STMN \cite{xiao2018STMN}& R101 &  & 80.5 & 2.49 & 13.2\\
         TROI \cite{gong2021temporalROI} &  R101 &  & 80.5 & 5.18 & 6.4\\
         SELSA \cite{wu2019sequence}& R101 &  & 80.5 & 3.15 & 10.6 \\
         OGEMN \cite{deng2019OGEMN} & R101+DCN &  & 81.6 & - & 8.9\\
         SparseVOD \cite{hashmi2022spatio}& R101 & \cmark & 81.9 & - & 14.4\\
         BoxMask \cite{hashmi2022boxmask} & R101 & \cmark & 83.2 & - & 6.1\\
         RDN \cite{deng2019RDN}& R101 &  & 83.8 & - & -\\
         HVRNet \cite{han2020mining}& R101 &  & 83.8 & -& -\\
         TF-Blender \cite{cui2021tf}& R101 & \cmark & 83.8 & - & 4.9\\
         MEGA \cite{chen2020memory}& R101 &  & 84.5 & 6.34 & 5.3 \\\midrule
         TransVOD \cite{zhou2022transvod} & \multirow{2}{*}{R101} & \multirow{2}{*}{\cmark} & \multirow{2}{*}{81.9} & \multirow{2}{*}{-} & \multirow{2}{*}{32.3} \\
         (Def. DETR) & & & & \\         
         PTSEFormer \cite{wang2022ptseformer} & \multirow{2}{*}{R101} & \multirow{2}{*}{\cmark} & \multirow{2}{*}{88.1} & \multirow{2}{*}{-} & \multirow{2}{*}{-}\\
         (Def. DETR) & & & & \\
         TransVOD \cite{zhou2022transvod} & \multirow{2}{*}{SwinB} & \multirow{2}{*}{\cmark} & \multirow{2}{*}{90.1} & \multirow{2}{*}{-} & \multirow{2}{*}{14.9} \\
         (Def. DETR) & & & & \\
         \midrule
         Ours (Faster RCNN) & R101 & \cmark & 87.2 & 0.78 & \textbf{39.6}\\
         Ours (Def. DETR) & R101 & \cmark & 87.9 & -&36.4 \\
         Ours (Def. DETR) & SwinB & \cmark & \textbf{91.3} & -&18.1 \\
    \bottomrule
    \end{tabular}}
    \caption{ \small
    \textbf{[C1]: Experiments on ImageNet VID.}
    We indicate the methods without video-level post-processing with a \cmark. 
    `No Post-proc.' means no post-processing. 
    R101 here is ResNet-101. 
    The runtime is measured on a NVIDIA GTX 1080 Ti. 
    Our method has the best performance and fastest runtime among all the methods using two-stage detectors (e.g. Faster RCNN).
    With a stronger detector and backbone, our method exceeds state-of-the-art.
    }
    \label{tab:ImageNetVID}
\end{table}

\medskip\noindent\textbf{Comparison on different motion speeds.} 
We evaluate across different motion speeds in \tab{ImageNetVID_speed}.
The category of object motion speeds in ImageNet VID follows FGFA \cite{zhu2017flow}.
Our method improves mAP on slow and medium motion speeds and achieves comparable results to the previous best method on fast motion speed, which shows the effectiveness of our method across different motion speeds.
\begin{table}[]
    \centering
    \resizebox{1\linewidth}{!}{%
    \begin{tabular}{l@{\hskip 0.1in}c@{\hskip 0.1in}c@{\hskip 0.1in}c@{\hskip 0.1in}r@{\hskip 0.1in}r}
    \toprule
        Methods & Backbone & mAP (\%)  & mAP (\%) & mAP (\%) & mAP (\%) \\  
        & & & (slow) & (medium) & (fast) \\ \midrule
         FGFA \cite{zhu2017flow} & R101 & 78.4 & 83.5 & 75.8 & 57.6 \\
         MANet \cite{wang2018fully} & R101 & 80.3 & 86.9 & 76.8 & 56.7 \\
         SELSA \cite{wu2019sequence}& R101 & 80.5 & 86.9 & 78.9 & 61.4 \\
         OGEMN \cite{deng2019OGEMN} & R101+DCN & 81.6 & 86.2 & 78.7 & 61.1\\
         HVRNet \cite{han2020mining} & R101 & 83.8 & 88.7 & 82.3 & \textbf{66.6} \\
         IFFNet \cite{jin2022feature} & R101 & 79.7 & 87.5 & 78.7 & 60.6 \\\midrule
         Ours (Faster RCNN) & R101 & \textbf{87.2} & \textbf{92.2} & \textbf{86.1} & 66.5\\
    \bottomrule
    \end{tabular}}
    \caption{\small\textbf{[C1]: ImageNet VID across different motion speeds}. 
    Our method improves mAP on different motion speeds.
    }
    \label{tab:ImageNetVID_speed}
\end{table}

\medskip\noindent\textbf{[C2]: Experiments on EPIC KITCHENS-55.} 
In EPIC KITCHENS-55, each frame contains avg/max 1.7/9 objects, which is more challenging compared to ImageNet VID. 
The Epic Kitchens video object detection task consists of 32 different kitchens and 290 classes. 
The training set has 272 video sequences captured in 28 kitchens. 
For evaluation, 106 sequences collected in the same 28 kitchens (S1) and 54 sequences collected in 4 other unseen kitchens (S2) are used. 
We use a prediction trajectory length of $T{=}4$ and evaluate for two IoU thresholds of 0.5 and 0.75.
As summarized in \tab{EPIC-kitchens}, our method is more accurate than previous state-of-the-art methods for both Seen/Unseen splits.
Our method is applicable to complex video detection tasks.

\begin{table}
    \centering
    \resizebox{\linewidth}{!}{%
    \begin{tabular}{l@{\hskip -0.05in}r@{\hskip 0.1in}r r@{\hskip 0.1in}r}
    \toprule
          & \multicolumn{2}{c}{S1} &  \multicolumn{2}{c}{S2}\\ \cmidrule(lr){2-3}\cmidrule(lr){4-5}
         Methods & mAP@.5 & mAP@.75 & mAP@.5 & mAP@.75 \\  \midrule
         EPIC \cite{damen2018scaling}& 34.2 & 8.5 &  32.0 & 7.9\\
         Faster-RCNN \cite{wu2019sequence}& 36.6  & 9.9 &  31.9 & 7.4\\
         SELSA \cite{wu2019sequence}& 37.9 & 9.8  & 34.8 & 8.1\\
         SELSA-ReIm + TROI \cite{gong2021temporalROI} &  42.2 & - & 39.6 & - \\
         BoxMask \cite{hashmi2022boxmask} & 44.3 & 18.5 &  41.3  & 15.7 \\\midrule
         Ours (Faster RCNN) & \textbf{44.9} & \textbf{18.7} &  \textbf{41.7} & \textbf{16.0}\\
    \bottomrule 
    \end{tabular}}
    \caption{ \small
    \textbf{[C2]: Experiments on EPIC KITCHENS-55.}
    S1 and S2 represent Seen and Unseen splits, respectively. 
    Our method achieves promising results for both test sets and IoU thresholds.}
    \label{tab:EPIC-kitchens}
\end{table}

\subsection{Sparsely annotated videos} 
\noindent\textbf{[S1] Sparsely annotated YouTube-BoundingBoxes.}
The YouTube-BoundingBoxes dataset \cite{real2017youtube} has sparse annotations: the video frame rate is 30 fps, and on average it only has annotations at 1 fps. 
We compare to the Faster RCNN \cite{ren2015faster} on YouTube-BoundingBoxes in \tab{YouTube-BoundingBoxes}, when using a Faster RCNN detector in our method.
The inputs are keyframes sampled with a step 60 and 30, which is every 60 frames and 30 frames in the video, respectively.
During training, for Faster RCNN we use the labels at the keyframes, while for our method we use labels with either a step of 30 or 1. 
For a label step of 30 and a keyframe step of 60, we use 2$\times$ more labels than input frames, while for a label step of 1 and keyframe step of 30, we use 30$\times$ more labels than frames. 
Since the video annotations are sparse, we do not have labels at every frame, therefore for a label step of 1, we use our sparse annotation loss, $L_\text{traj}^\text{(sa)}$.
In the $L_\text{traj}^\text{(sa)}$ we define the box trajectories to be linearly interpolated pseudo trajectories.
For keyframes sampled with a step of 60, our method has higher accuracy than Faster RCNN by using $2\times$ more labels and processing the same input keyframes.
With input keyframes every 30 frames, our method achieves a 1.1\% higher mAP than Faster RCNN while optimizing a simulated motion between these frames.

\begin{table}[]
    \centering
    \resizebox{.75\linewidth}{!}{%
    \begin{tabular}{l c c c}
    \toprule
        \multirow{2}{*}{Methods} & Keyframe & Label & \multirow{2}{*}{mAP (\%)} \\ 
        & step & step & \\ 
        \midrule
         Faster RCNN \cite{ren2015faster}                & 60 & 60 & 47.6\\ 
         Faster RCNN \cite{ren2015faster}                & 30 & 30 & 58.7\\ \midrule
         Ours $L_\text{traj}$ \vspace{2px}               & 60 & 30 & 51.3 \\ \cdashline{1-4}
         Ours $L_\text{traj}^{\text{(sa)}}$              & 30 & 1 & \textbf{59.8}\\
    \bottomrule
    \end{tabular}}
    \caption{\small\textbf{[S1] Sparsely annotated YouTube-BoundingBoxes.}
    mAP on the sparsely annotated YouTube-BoundingBoxes. 
    The sparse annotation loss $L_\text{traj}^{\text{(sa)}}$ using interpolated pseudo trajectory labels improves keyframe detection.
    }
    \label{tab:YouTube-BoundingBoxes}
\end{table}

\medskip\noindent\textbf{[S2] Results on Waymo Open Dataset.}
We are the first to perform video object detection on the Waymo Open dataset \cite{Ettinger_2021_ICCV}, and thus we can only compare to a static detector, Faster RCNN \cite{ren2015faster}.
We use a Faster RCNN detector as well in our method, and the sparse annotations loss with linearly interpolated pseudo-trajectories.
The Waymo Open dataset contains sparse object annotations at 1 fps.
Nonetheless, the results in \tab{waymo} show that we outperform Faster-RCNN. 

\begin{table}[]
    \centering
    \resizebox{0.6\linewidth}{!}{%
    \begin{tabular}{l@{\hskip 0.1in}c@{\hskip 0.1in}r}
    \toprule
    Methods & AP/L1 (\%) & AP/L2 (\%) \\
    \midrule
    Faster RCNN & 55.66 & 49.63 \\
    \midrule
    Ours $L_\text{traj}^{\text{(sa)}}$  & 64.53 & 59.28 \\
    \bottomrule 
    \end{tabular}}
    \caption{\small\textbf{[S2]: Experiments on the Waymo Open dataset}. 
    We report average precision over the $L1$ and $L2$ instance difficulty levels.
    Even when learning from object annotations at 1 fps, our method using interpolated pseudo-trajectories outperforms the Faster RCNN baseline.
    }
    \label{tab:waymo}
\end{table}

%% file: Paper/conclusion.tex
\subsection{Method limitations}
\begin{figure}[t]
    \centering
    \begin{tabular}{cc}
        \includegraphics[trim={5cm 2.5cm 5cm 2.5cm},clip,width=.45\linewidth]{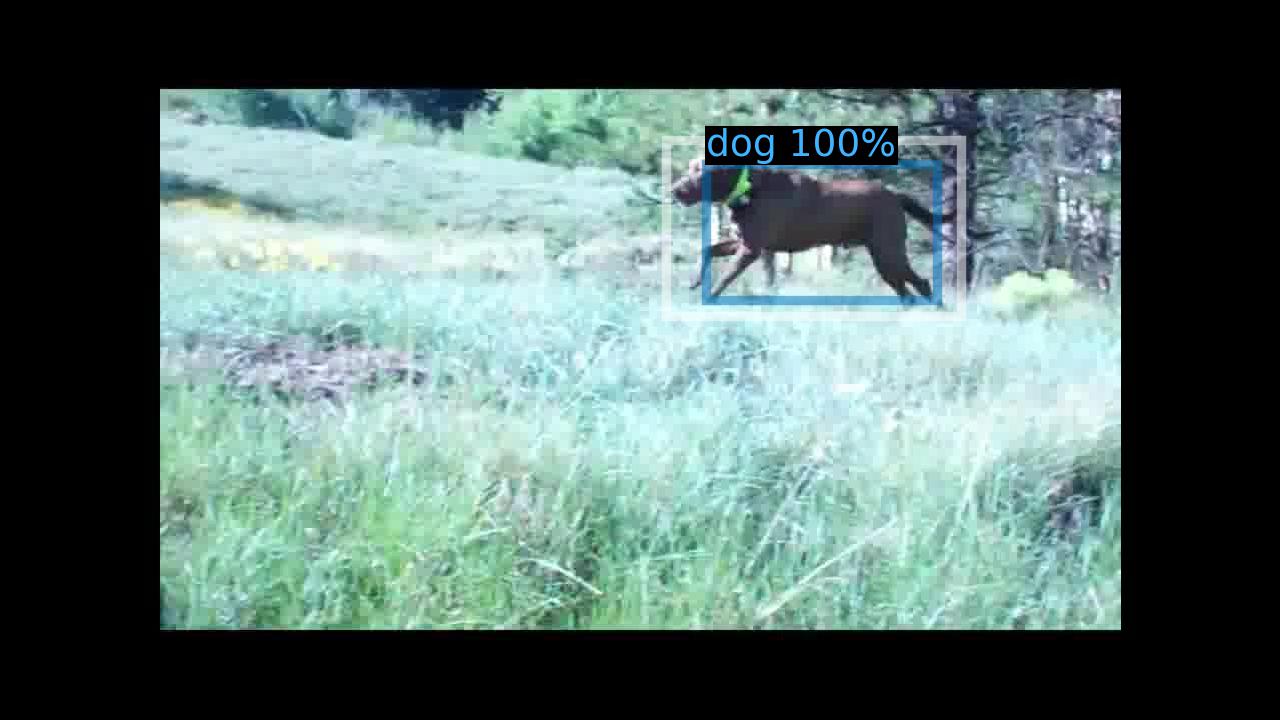} &
        \includegraphics[trim={5cm 5cm 5cm 0cm},clip,width=.45\linewidth]{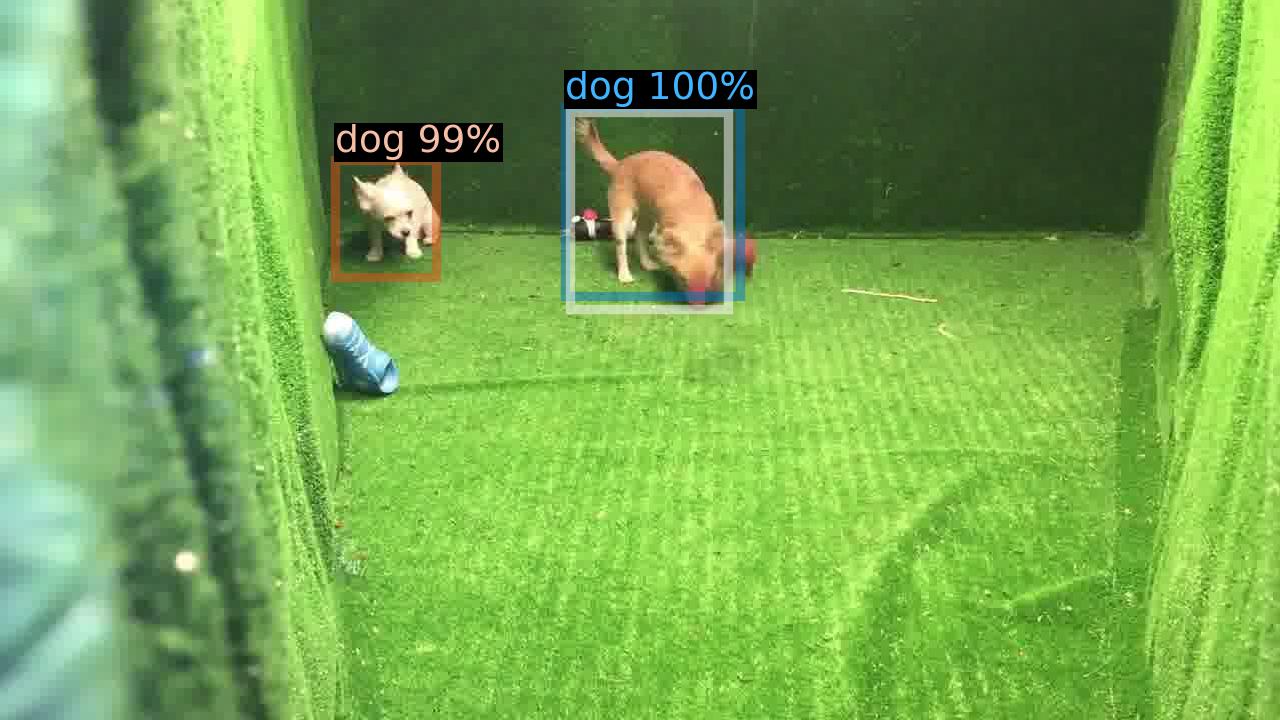} \\ 
        \includegraphics[trim={5cm 2.5cm 5cm 2.5cm},clip,width=.45\linewidth]{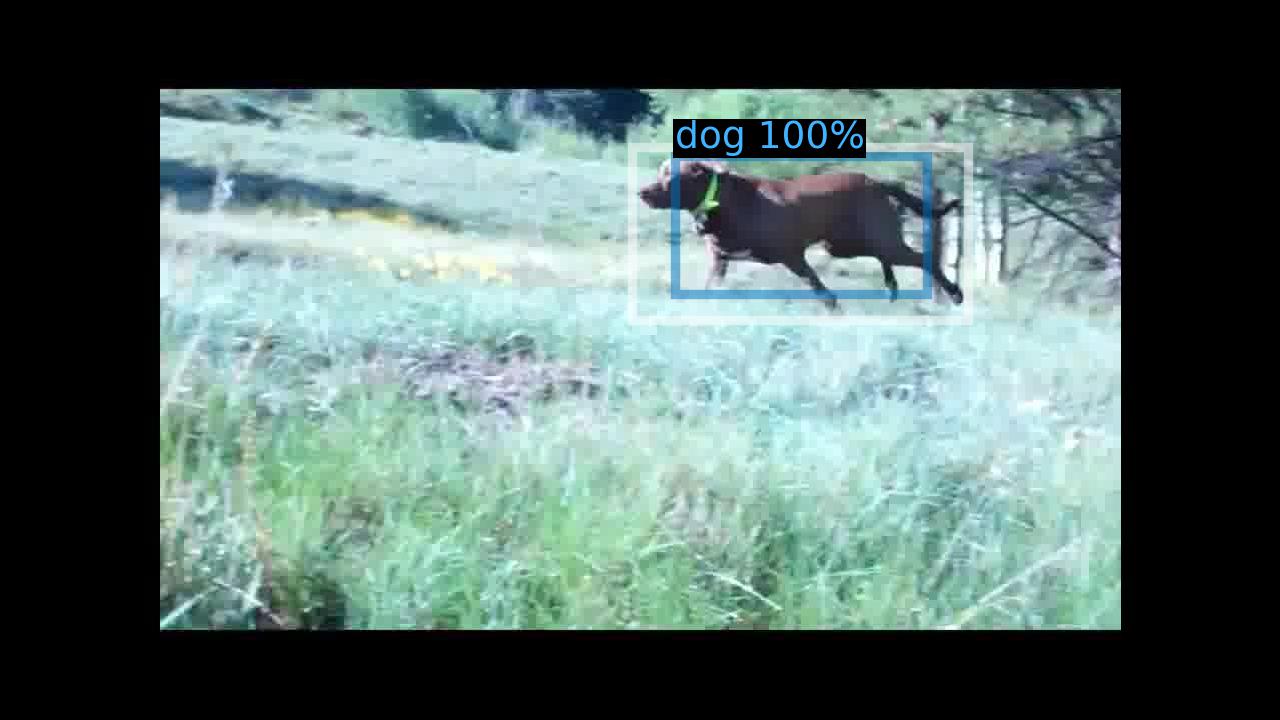} &
        \includegraphics[trim={5cm 5cm 5cm 0cm},clip,width=.45\linewidth]{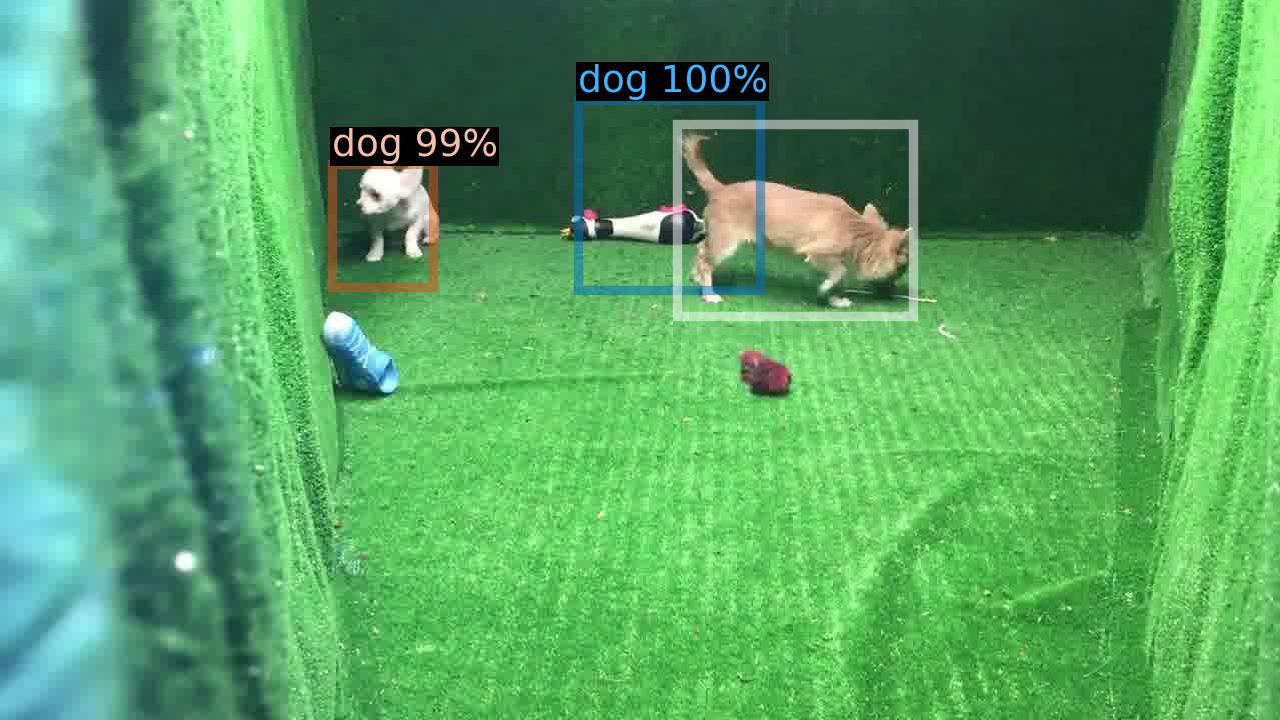} \\ 
        \includegraphics[trim={5cm 2.5cm 5cm 2.5cm},clip,width=.45\linewidth]{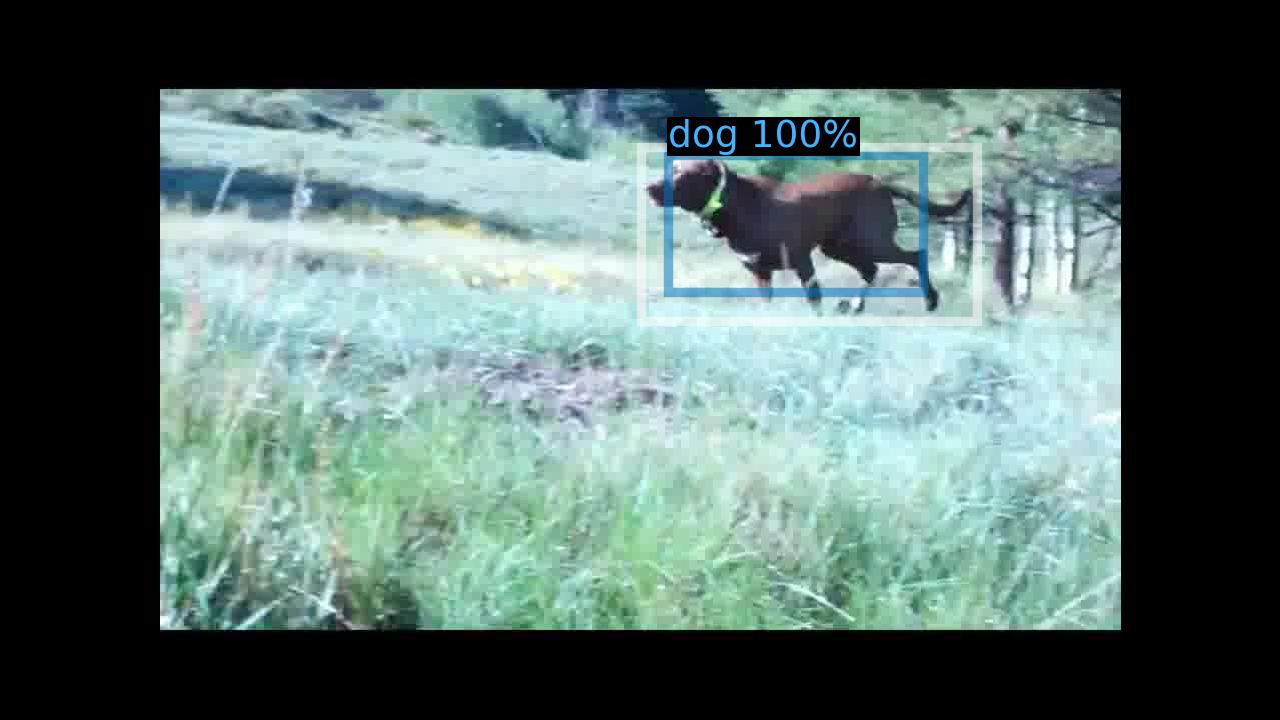} &
        \includegraphics[trim={5cm 5cm 5cm 0cm},clip,width=.45\linewidth]{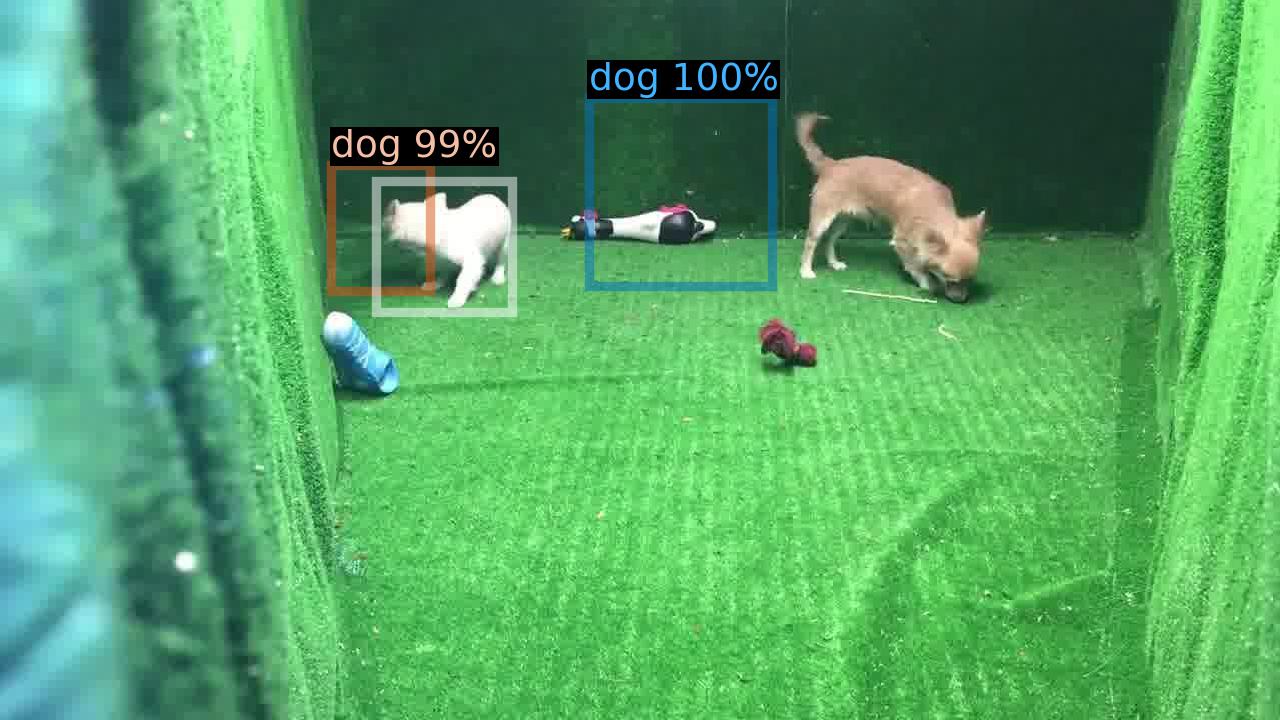} \\ 
        \includegraphics[trim={5cm 2.5cm 5cm 2.5cm},clip,width=.45\linewidth]{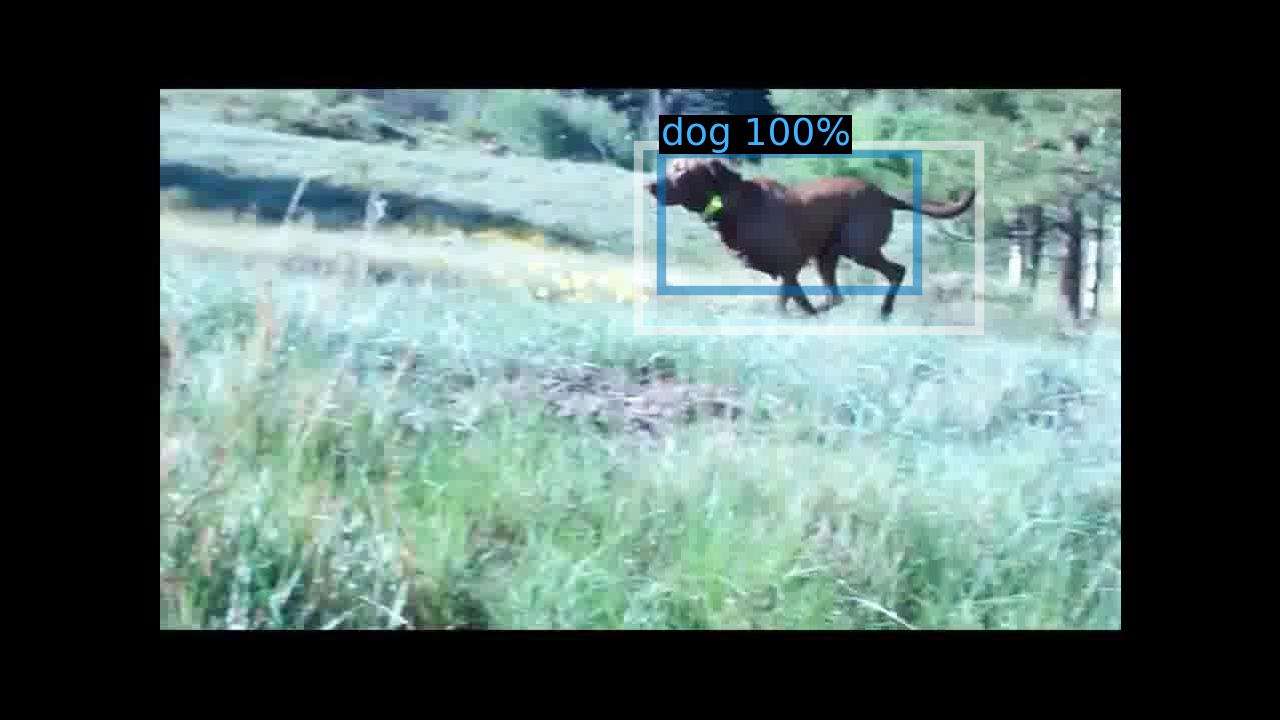} &
        \includegraphics[trim={5cm 5cm 5cm 0cm},clip,width=.45\linewidth]{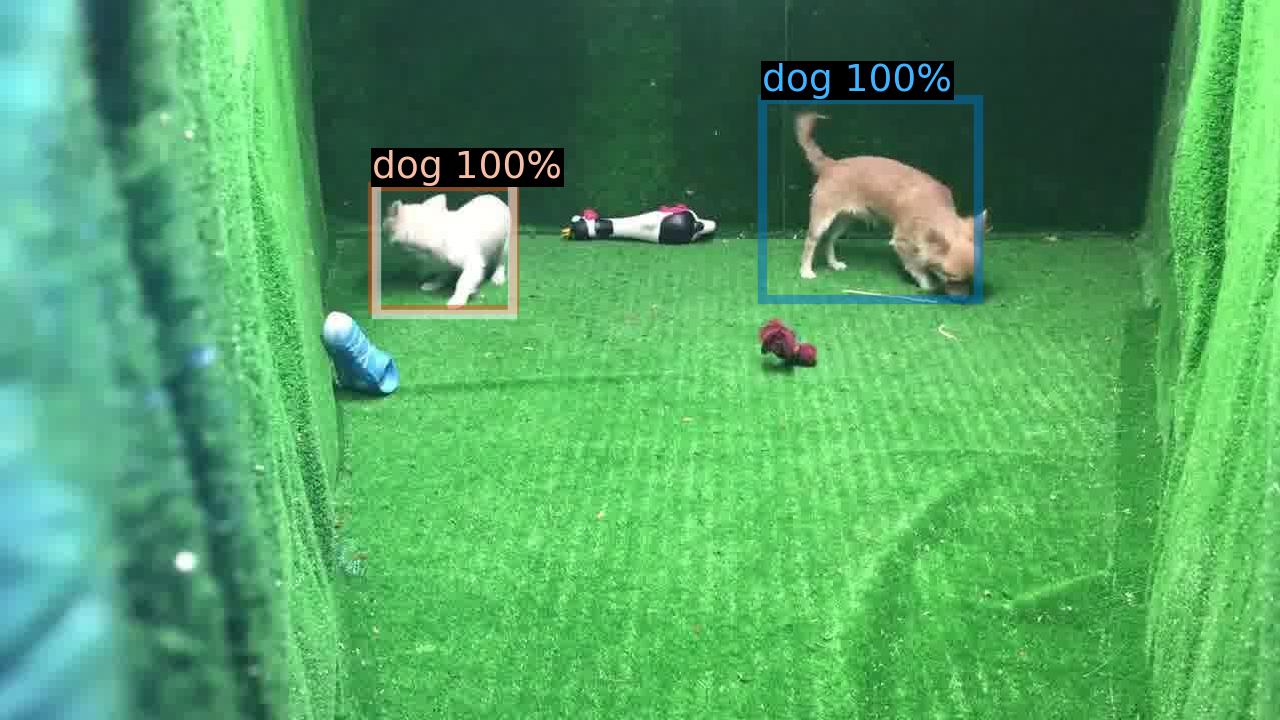} \\ 
        {\small(a) Success case} & {\small(b) Failure case} \\
    \end{tabular}   
    \caption{\small\textbf{Example predictions.} 
        White boxes are the ground truth, blue boxes are the predictions. 
        We show one success case in (a) a `dog' running. 
        In (b) the failure case shows a `dog' suddenly changing direction.
        We miss the detection in the second and third frames, yet we recover at the keyframe detection.
        When the motion has large variations, is unpredictable, and objects enter and leave the frame, our method fails.
     }
    \label{fig:examples}
\end{figure}

Despite our method\rq s successful predictions as in \fig{examples}(a), we also identify several limitations.
In practice, multiple motion patterns can be associated with the same appearance: e.g people can walk or jump.
In this case, our method may fail to predict the correct object locations.
Another failure case is if objects appear or disappear in the middle of a trajectory. 
In these cases, we will either miss the objects or over-predict.
Yet another limitation of our method is the assumption that the motion changes smoothly. 
If the object trajectories have large variations in a short time because of the video frame rate, our trajectory network may not be able to learn this.
One such example is the dog in \fig{examples}(b) which suddenly changes its moving direction. 
Even if we miss some detections of intermediate frames, as shown in \fig{examples}(b), our method can recover at the next keyframe detection. 
While these limitations exist, they only influence our method minimally, since we miss only a few hundred milliseconds, which is reflected in our state-of-the-art video object detection results. 

\section{Conclusion}
We propose a method to efficiently detect objects in videos by predicting their future locations from a static input keyframe and the ground truth locations of all frames.
Our method associates appearance with motion. 
Different motion contexts have different appearances, thus we can model various motion patterns from the keyframes with different appearances.
Because we predict the future object locations over multiple frames, we do not need to process every frame of the video, but only a subset of the keyframes, which makes our method efficient.
Moreover, by learning to predict object trajectories we improve the object detection accuracy when compared to the state-of-the-art on multiple datasets.
Finally, by using pseudo object trajectories defined by smooth continuous functions, we can improve object detection accuracy on sparsely annotated videos.

{\small\noindent\textbf{Acknowledgments.}
This work is part of the research program Efficient Deep Learning (EDL), which is (partly) financed by the Dutch Research Council (NWO). 
Fatemeh Karimi Nejadasl is financed by the University of Amsterdam Data Science Centre.}

%% file: Paper/supplementary.tex
\onecolumn
\begin{center}
    \textbf{\large Supplementary Material}
\end{center}
\appendix

\begin{multicols}{2}
\section{Derivation for loss equivalence}
In this appendix section, we provide the derivation to show that if we would predict bounding boxes $B_{t+l}$ in the trajectory network, instead of offsets between pairs of bounding boxes $\delta_{t+l}$, \eq{lsum} would reduce to \eq{loss} and the temporal ordering would not be enforced.

If we predict bounding boxes $B_{t+l}$ and use $\delta_{t+k} = (B_{t+k} - B_{t+k-1}), k\in\{1, .., l\}$,  the sum $\sum_{k=1}^{l} \delta_{t+k}$ can be rewritten as follows:
\begin{eqnarray*}
     \sum_{k=1}^{l} \delta_{t+k} & = & \sum_{k=1}^{l} (B_{t+k} - B_{t+k-1}) , \\ \nonumber
                            & = & \sum_{k=1}^{l} (B_{t+k}) - \sum_{k=1}^{l} (B_{t+k-1}) , \\ \nonumber
                            & = & \sum_{k=1}^{l} (B_{t+k}) - \sum_{k=0}^{l-1} (B_{t+k}), \\ \nonumber
                            & = & \sum_{k=1}^{l-1} (B_{t+k}) + B_{t+l} - \sum_{k=1}^{l-1}(B_{t+k}) - B_{t}, \\ \nonumber
                            & = & B_{t+l}  - B_{t}.
\end{eqnarray*}
And we fill the above in \eq{lsum}. Then we have,
\begin{eqnarray*}
    L_{\sum}(B^*, \overleftrightarrow{\mathbb{T}_t}) & = & \sum_{l=0}^T {\mathcal{L}_1}\left( \left( {B}^*_{t+l} - B_t \right) - \sum_{k=1}^{l} \delta_{t+k} \right), \\ \nonumber
                               & = & \sum_{l=0}^T {\mathcal{L}_1}\left( {B}^*_{t+l} - B_t - \sum_{k=1}^{l} \delta_{t+k} \right), \\ \nonumber
                               & = & \sum_{l=0}^T {\mathcal{L}_1}\left( {B}^*_{t+l} - B_t - B_{t+l} + B_{t} \right), \\ \nonumber
                               & = & \sum_{l=0}^T {\mathcal{L}_1}\left( {B}^*_{t+l} - B_{t+l} \right).
\end{eqnarray*}
which is the same as \eq{loss}:
\begin{align*}
    L_{\text{bag}}(B^*, \{B_t, .., B_{t+T}\}) = \sum_{l=0}^{T} \mathcal{L}_1({B}^*_{t+l} - {B}_{t+l}).
\end{align*}

\section{Details for the \textit{Simulated smooth motion}}
\begin{figure}[H]
    \centering
    \includegraphics[width=1.\linewidth]{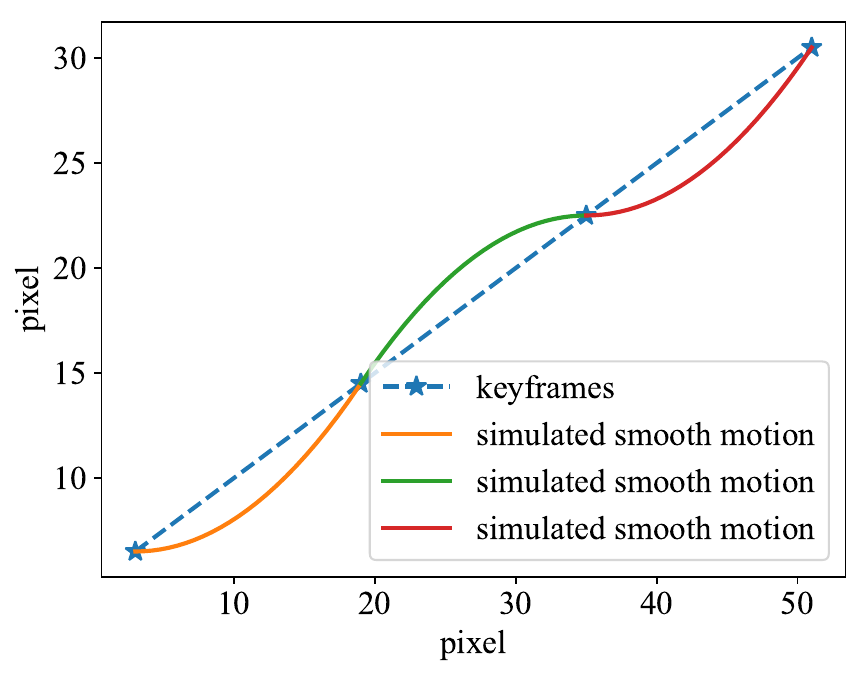}
    \caption{ \small
     An example of simulated smooth motion generated by parabola functions. The parabola represents the trajectory of intermediate digit locations between every two keyframes. The simulated motion is smooth and continuous.
    }
    \label{fig:parabola}
\end{figure}

In this section, we describe how we create the \textit{Simulated smooth motion}: bounding boxes move between keyframes according to a smooth parabola, and the change of width and height is linearly interpolated.
Given the center points of two keyframe digits $(x_{t}, y_{t})$ and $(x_{t+T}, y_{t+T})$, we choose the focus $F=(0, f), f=8$ for the parabola, then the parabola can be written as,
\begin{align}
    y = \frac{1}{4f}x^{2}-\frac{v_{1}}{2f}x+\frac{v_{1}^{2}}{4f} +v_{2},
\end{align}
where the vertex is $V=(v_{1}, v_{2})$.
By filling in $(x_{t}, y_{t})$ and $(x_{t+T}, y_{t+T})$, we can get the value of $v_{1}, v_{2}$.
For every pairwise neighbouring keyframes, we can have a parabola that acts as a simulated smooth trajectory for intermediate locations of digits.
Here we show an example of having four keyframes and the simulated smooth motion as a parabola in \fig{parabola}.
Because the digits move linearly in MovingDigits dataset, the digits of the keyframes stay on a linear line.
We choose the focus of every second parabola sequence to be $F=(0, -8)$ to make all the parabola trajectories smoothly connected.

\clearpage
\end{multicols}